\documentclass[11pt]{article}

\usepackage[final]{acl}

\usepackage{times}
\usepackage{latexsym}

\usepackage[T1]{fontenc}
\usepackage{microtype}
\usepackage{inconsolata}
\usepackage{graphicx}
\usepackage{subcaption}
\usepackage{booktabs}
\usepackage{amsmath}
\usepackage{amssymb}
\usepackage{mathtools}
\usepackage{amsthm}
\usepackage{multirow}

\theoremstyle{plain}

\theoremstyle{definition}

\theoremstyle{remark}

\usepackage[textsize=tiny]{todonotes}

\usepackage{tabularx}
\usepackage[table]{xcolor}
\usepackage{makecell}

\definecolor{bg_red}{RGB}{253, 235, 235}
\definecolor{bg_blue}{RGB}{235, 245, 253}
\definecolor{row-highlight}{RGB}{220, 230, 242}

\newcolumntype{Y}{>{\centering\arraybackslash}X}
\newcolumntype{R}{>{\columncolor{bg_red}}Y}
\newcolumntype{B}{>{\columncolor{bg_blue}}Y}

\usepackage[capitalize,noabbrev]{cleveref}

\usepackage{xcolor}
\usepackage{tcolorbox}
\usepackage{listings}
\tcbuselibrary{listings, skins, breakable}

\definecolor{bg_color}{RGB}{250, 250, 250}
\definecolor{frame_color}{RGB}{80, 80, 80}

\definecolor{budget_col}{RGB}{197, 90, 17}
\definecolor{think_col}{RGB}{68, 114, 196}
\definecolor{search_col}{RGB}{0, 176, 240}
\definecolor{info_col}{RGB}{191, 144, 0}
\definecolor{ans_col}{RGB}{192, 0, 0}

\lstdefinelanguage{PromptLang}{
    sensitive=true,
    literate={Question:}{{\textbf{Question:}}}9
             {Ground}{{\textbf{Ground}}}6
             {Truth:}{{\textbf{Truth:}}}6
             {<budget>}{{\textcolor{budget_col}{<budget>}}}1
             {</budget>}{{\textcolor{budget_col}{</budget>}}}1
             {<think>}{{\textcolor{think_col}{<think>}}}1
             {</think>}{{\textcolor{think_col}{</think>}}}1
             {<search>}{{\textcolor{search_col}{<search>}}}1
             {</search>}{{\textcolor{search_col}{</search>}}}1
             {<information>}{{\textcolor{info_col}{<information>}}}1
             {</information>}{{\textcolor{info_col}{</information>}}}1
             {<answer>}{{\textcolor{ans_col}{<answer>}}}1
             {</answer>}{{\textcolor{ans_col}{</answer>}}}1,
    basicstyle=\ttfamily\footnotesize\color{black},
    breaklines=true,
    breakatwhitespace=true,
    breakindent=0pt,
    columns=fullflexible,
    keepspaces=true,
    showstringspaces=false,
}

\newtcblisting{promptbox}[1][]{
    enhanced,
    breakable,
    colback=bg_color,
    colframe=frame_color,
    boxrule=0.6pt,
    arc=2pt,
    left=4pt,
    right=4pt,
    top=4pt,
    bottom=4pt,
    listing only,
    listing options={
        language=PromptLang,
    },
    title=\textbf{Training Prompt},
    fonttitle=\sffamily\small,
    #1
}

\newtcblisting{caseboxin}[1][]{
    enhanced,
    breakable,
    colback=bg_color,
    colframe=frame_color,
    boxrule=0.6pt,
    arc=2pt,
    left=4pt,
    right=4pt,
    top=4pt,
    bottom=4pt,
    listing only,
    listing options={
        language=PromptLang,
    },
    title=\textbf{Case 1 (Search-R1 Budget=6)},
    fonttitle=\sffamily\small,
    #1
}

\newtcblisting{caseboxex}[1][]{
    enhanced,
    breakable,
    colback=bg_color,
    colframe=frame_color,
    boxrule=0.6pt,
    arc=2pt,
    left=4pt,
    right=4pt,
    top=4pt,
    bottom=4pt,
    listing only,
    listing options={
        language=PromptLang,
    },
    title=\textbf{Case 2 (AnySearch Budget=6)},
    fonttitle=\sffamily\small,
    #1
}

\newtcblisting{caseboxinex}[1][]{
    enhanced,
    breakable,
    colback=bg_color,
    colframe=frame_color,
    boxrule=0.6pt,
    arc=2pt,
    left=4pt,
    right=4pt,
    top=4pt,
    bottom=4pt,
    listing only,
    listing options={
        language=PromptLang,
    },
    title=\textbf{Case 3 (Search-R1 Budget=3)},
    fonttitle=\sffamily\small,
    #1
}

\newtcblisting{caseboxinex2}[1][]{
    enhanced,
    breakable,
    colback=bg_color,
    colframe=frame_color,
    boxrule=0.6pt,
    arc=2pt,
    left=4pt,
    right=4pt,
    top=4pt,
    bottom=4pt,
    listing only,
    listing options={
        language=PromptLang,
    },
    title=\textbf{Case 4 (AnySearch Budget=3)},
    fonttitle=\sffamily\small,
    #1
}

\newtcblisting{caseboxinex3}[1][]{
    enhanced,
    breakable,
    colback=bg_color,
    colframe=frame_color,
    boxrule=0.6pt,
    arc=2pt,
    left=4pt,
    right=4pt,
    top=4pt,
    bottom=4pt,
    listing only,
    listing options={
        language=PromptLang,
    },
    title=\textbf{Case 5 (AnySearch with Scaffold Budget=6)},
    fonttitle=\sffamily\small,
    #1
}

\title{One Policy, Any Budget: Internalizing Budget-Aware Search via Reinforcement Learning}

\author{
  \textbf{Xiaowei Sun}\textsuperscript{1},
  \textbf{Jin Li}\textsuperscript{2},
  \textbf{Yili Hong}\textsuperscript{1},
  \textbf{Yikun Fu}\textsuperscript{3},
  \textbf{Yanghua Xiao}\textsuperscript{1}\thanks{Corresponding author.}
  \\
  \textsuperscript{1}College of Computer Science and Artificial Intelligence, Fudan University \\
  \textsuperscript{2}College of Software Engineering, Southeast University \\
  \textsuperscript{3}School of Artificial Intelligence, Shanghai Jiao Tong University \\
  \{xwsun24, ylhong24\}@m.fudan.edu.cn, \\
  jin\_li@seu.edu.cn, fuyikun123456@sjtu.edu.cn, \\
  shawyh@fudan.edu.cn
}

\begin{document}
\maketitle

\begin{abstract}

While reinforcement learning has enabled LLM-based search agents to invoke external tools, existing methods train under fixed budgets and cannot adapt when constraints vary at deployment. We propose \textbf{AnySearch}, a framework that enables \textbf{a single policy} to perform budget-aware search under \textbf{any budget} constraint through a training scaffold and curriculum reinforcement learning. In the first phase, we train the agent with explicit budget state injection and structured reasoning prompts that guide efficient allocation under linearly decaying budgets. In the second phase, the scaffold is removed and the agent learns to operate autonomously under adaptively sampled budget constraints, matching inference conditions. Both phases are optimized with a composite reward that couples answer accuracy with budget efficiency through absolute and relative signals, where an adaptive weight amplifies the efficiency signal for high-accuracy queries and attenuates it for low-accuracy ones. Extensive experiments on seven general and multi-hop QA benchmarks show that our method outperforms baselines across all budget scales, generalizes to unseen constraints beyond the training range, and achieves superior tool productivity without excessive token overhead. Our code is available at \url{https://github.com/xwsun01/AnySearch}.

\end{abstract}

\section{Introduction}

\begin{figure}[t]
\centering
\includegraphics[width=\columnwidth]{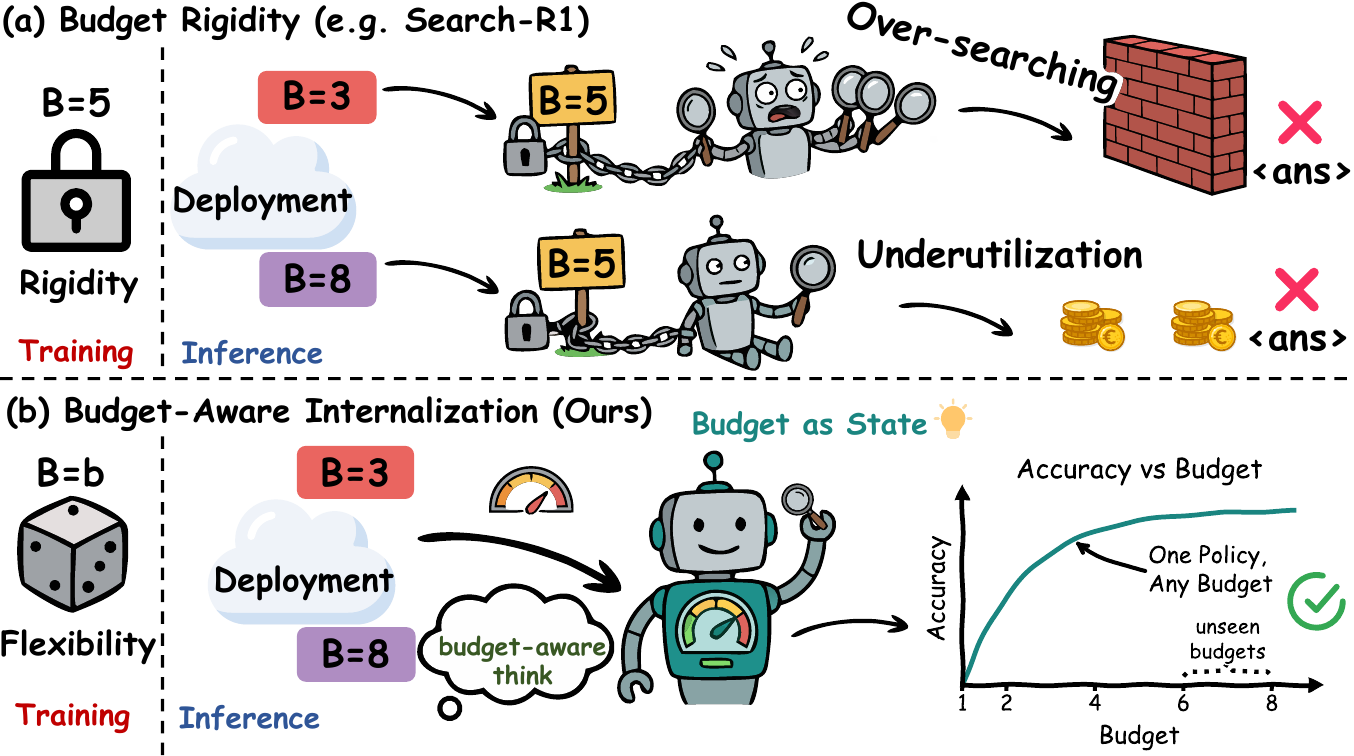}
\caption{Comparison of search behaviors under varying budgets. Existing methods trained under fixed budgets cannot adapt their search strategy when the budget changes, while our method adjusts its allocation proportionally to the available budget.}
\label{fig:cover}
\end{figure}

Augmenting Large Language Models (LLMs) with external tools has become a fundamental approach to compensate for the limitations of static parametric knowledge~\cite{qin2024tool,qu2025tool,schick2023toolformer}. Reinforcement learning (RL) further enables search agents to autonomously decide when and what to search through multi-turn interaction~\cite{jin2025search,song2025r1,chen2025learning,yao2022react}. However, existing RL-based search agents are predominantly trained under fixed or unconstrained budgets, producing policies that cannot adapt when resource availability changes at deployment~\cite{jin2025search,wang2025stepsearch}.

In practice, search budgets vary widely across diverse deployment scenarios, from latency-critical applications requiring minimal external search calls to deep research tasks permitting extensive exploration~\cite{jeong2024adaptive,zheng2025deepresearcher}. An ideal search agent should operate effectively under any budget constraint with a single policy, rather than requiring retraining or separate models for different resource regimes (\cref{fig:cover}). This requires more than simply capping tool calls at a threshold. The agent must learn to allocate limited budgets effectively, deciding which queries merit external search and which can be resolved through reasoning~\cite{asai2024self,jiang2023active,xi2025survey}. 

Prior approaches fall short of this goal. Methods that reduce redundant tool calls via reward shaping~\cite{wang2025acting,huang2025reinforced} lack an explicit budget interface and cannot adapt to arbitrary constraints specified at inference. BATS~\cite{liu2025budget} introduces budget state tracking, but relies on an external tracker that must persist at inference, leaving budget-aware allocation as an external dependency rather than an internalized capability of the policy itself.

Accordingly, we propose \textbf{AnySearch} that internalizes budget-aware search into the policy, so that at inference the agent requires only a total budget specification and autonomously handles all allocation decisions. Our key insight is that budget-aware behavior can be scaffolded during training and internalized into the policy, eliminating external dependencies at inference~\cite{wood1976role,yu2024distilling}.

Specifically, we design a \textbf{training scaffold} that provides explicit budget state tracking and structured reasoning prompts, guiding the agent toward budget-aware decision patterns. To internalize these patterns, we employ a \textbf{two-phase curriculum} that progressively removes the scaffold. Phase~I trains with the full scaffold under linearly decaying budgets, allowing the agent to learn search behaviors with explicit guidance. Phase~II removes the scaffold entirely and applies adaptive budget sampling that focuses training on weak budget levels while maintaining coverage through uniform smoothing. Since Phase~II matches inference conditions exactly, the training-inference gap is eliminated. The learning signal is provided by a \textbf{composite reward} that couples answer accuracy with budget efficiency through absolute and relative signals, where an adaptive weight dynamically adjusts the balance based on per-query difficulty, preventing the agent from sacrificing correctness on hard queries.

We evaluate on seven benchmarks across three backbone models. Our method consistently outperforms baselines at all tested budget levels, generalizes to budget constraints unseen during training, achieves the highest tool productivity, and produces the lowest total token consumption. Ablation studies confirm that the scaffold is successfully internalized after removal, and that each component of the curriculum and reward design contributes to the final performance.

Our contributions are summarized as follows:
\begin{itemize}
\item We propose a framework that internalizes budget-aware search through a progressively removed training scaffold, eliminating external dependencies at inference.
\item We introduce a two-phase curriculum with sliding-window-based adaptive budget sampling and a composite reward with adaptive efficiency weighting to jointly optimize accuracy and efficiency.
\item Extensive experiments demonstrate consistent improvements, robust generalization to unseen budgets, and superior tool productivity. Ablations validate the internalization hypothesis and each component's necessity.
\end{itemize}

\section{Related Work}

\textbf{Budget-Aware Reasoning.}
Efficiently allocating finite computational resources has become central to LLM inference research~\cite{chen2026token}. For \textit{token-budget} constrained reasoning, BRPO~\cite{qi2025optimizing} and BudgetThinker~\cite{wen2025budgetthinker} train under varying token constraints via RL, but still require explicit budget signals at inference to control reasoning depth. For \textit{tool-budget} constrained agents, OTC-PO~\cite{wang2025acting} and IKEA~\cite{huang2025reinforced} reduce redundant tool calls via reward shaping, yet lack an explicit budget interface and cannot adapt to arbitrary constraints. BATS~\cite{liu2025budget} introduces budget state tracking for tool-call scaling. However, it relies on an external tracker at inference and does not internalize budget-aware capability into the policy model. In contrast, our method internalizes budget-aware capability via RL, requiring only a total budget specification at inference and adapting to any budget without retraining.

\textbf{Agentic RL with Search Engines.}
Recent work trains LLMs to interleave reasoning with external search via RL~\cite{wei2026agentic,lin2025comprehensive,pati2025agentic,acharya2025agentic}. Search-R1~\cite{jin2025search}, R1-Searcher~\cite{song2025r1}, ReSearch~\cite{chen2025learning}, and DeepResearcher~\cite{zheng2025deepresearcher} optimize multi-turn interaction trajectories with outcome-based reward, learning when and what to search. To address the sparsity of the reward, StepSearch~\cite{wang2025stepsearch} and AutoRefine~\cite{shi2025search} introduce step-level supervision. Furthermore, ZeroSearch~\cite{sun2025zerosearch} and SSRL~\cite{fan2025ssrl} replace external search engines with LLM-simulated retrieval environments, reducing API costs while enabling scalable training. These methods focus on improving search effectiveness under fixed budgets, without explicitly modeling how to allocate limited search budgets or adapt to any constraint at inference. To our knowledge, no prior work trains search agents under dynamic tool-call budgets.

\section{Method}
\label{sec:method}

\begin{figure*}[t]
\centering
\includegraphics[width=\textwidth]{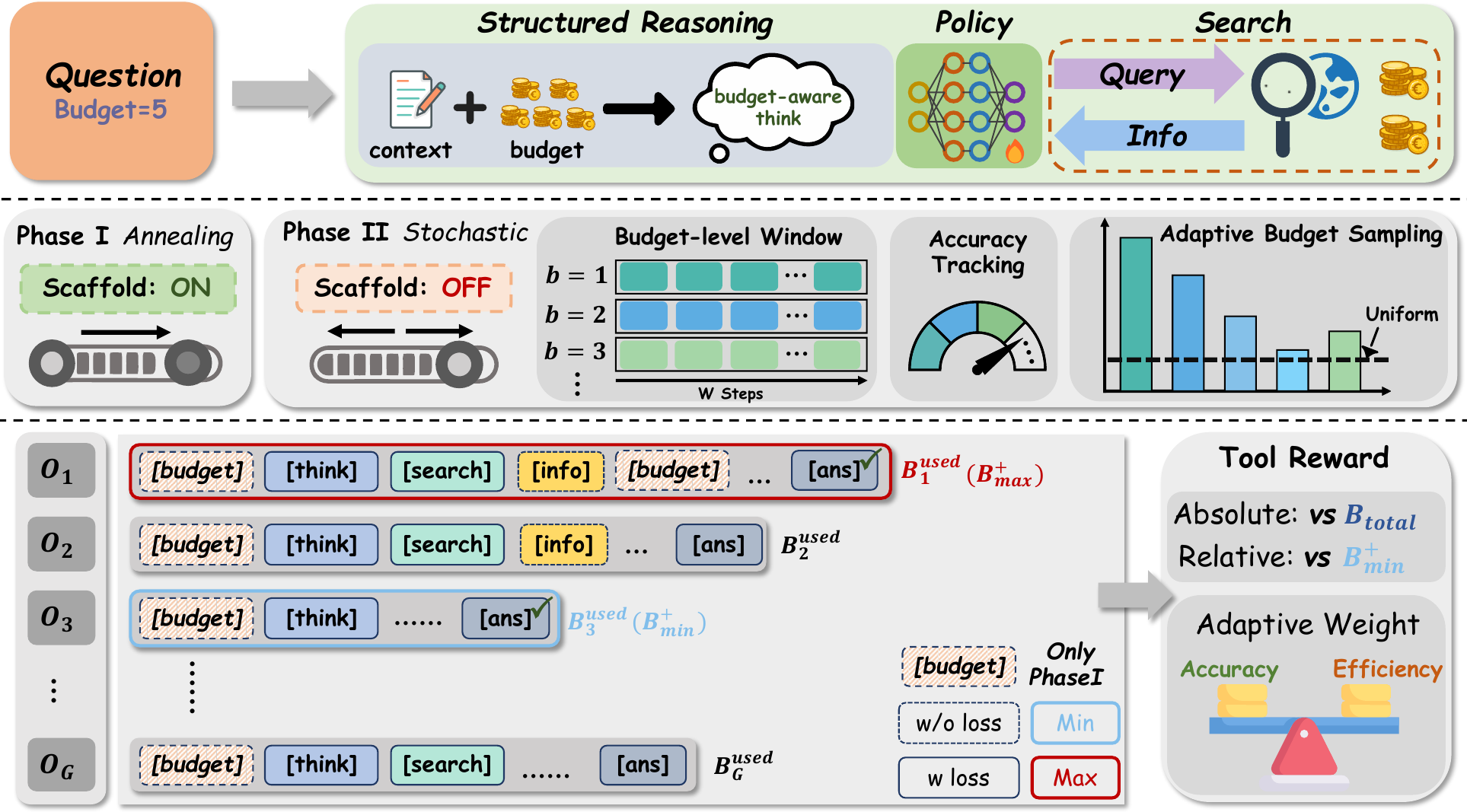}
\caption{Overview of our framework. Phase I trains the agent with a reasoning scaffold and linearly decaying budgets to establish budget-aware decision patterns. Phase II removes the scaffold and applies sliding-window-based adaptive budget sampling and a composite reward with adaptive efficiency weighting.}
\label{fig:overall}
\end{figure*}

This section describes how we internalize budget-aware search via RL. The method comprises task formulation (\cref{sec:task}), a training scaffold (\cref{sec:scaffold}), a two-phase curriculum (\cref{sec:curriculum}), and reward design (\cref{sec:reward}). The overall framework is illustrated in \cref{fig:overall}.

\subsection{Task Formulation}
\label{sec:task}

We formulate budget-aware agentic search as a resource-constrained sequential decision problem. Given a question $q$ and an assigned search budget $B \in \mathbb{Z}_{\geq 0}$, the agent must produce an accurate answer while consuming at most $B$ search calls.

\paragraph{Agent Loop.}
At each step $t$, the agent observes its current state $s_t = (h_t, b_t)$, where $h_t$ denotes the accumulated context and $b_t$ is the remaining budget. Then it takes one of actions $a_t$: \textbf{Reason} ($\mathcal{A}_{think}$) performs reasoning within \texttt{<think>} tags. \textbf{Search} ($\mathcal{A}_{tool}$) calls the search engine via \texttt{<search>} tag and consumes one unit of budget. \textbf{Answer} ($\mathcal{A}_{ans}$) outputs the final answer via \texttt{<answer>} tag and terminates the episode.

The budget evolves deterministically:
\begin{equation}
    b_{t+1} = b_t - \mathbb{I}(a_t \in \mathcal{A}_{tool})
\end{equation}
When the budget is exhausted ($b_t = 0$), the agent can no longer search but may still reason before outputting the final answer. The core challenge lies not only in respecting the budget ceiling, but in learning to allocate it effectively. We refer to this as the budget-aware search capability and aim to internalize it via reinforcement learning. Formally, the agent learns a policy $\pi_\theta(a_t | s_t)$ that generates a trajectory $\tau = (s_0, a_0, \ldots, s_T)$, the goal is to maximize the expected reward:
\begin{equation}
    \max_\theta \mathbb{E}_{\tau \sim \pi_\theta} \left[ R(\tau) \right] \quad \text{s.t.} \quad \sum_{t} \mathbb{I}(a_t \in \mathcal{A}_{tool}) \leq B
\end{equation}
where $R(\tau)$ is the composite reward detailed in \cref{sec:reward}. The budget is defined over external search calls, as each call is orders of magnitude more expensive than equivalent inference tokens and the returned documents dominate total token consumption (Appendix~\ref{app:cost}).

\subsection{Budget-Aware Training Scaffold}
\label{sec:scaffold}

To internalize budget-aware search capabilities, we employ a training scaffold that provides explicit budget state tracking and structured reasoning prompts. The entire scaffold is removed at Phase~II of training (\cref{sec:curriculum}) and remains absent at inference.

\paragraph{Budget State Injection.}
Motivated by ~\cite{liu2025budget}, we explicitly inject the budget state at the onset of each turn, containing the remaining, used and total budget:
\begin{center}
    \small\texttt{<budget> remaining=R; used=U; total=T </budget>}
\end{center}
This makes budget state fully observable, enabling the agent to condition its search decisions on resource availability at each step.

\paragraph{Structured Reasoning Prompts.}
We prompt the agent to conduct a dual-factor analysis within the \texttt{<think>} block~\cite{wei2022chain}: (1) \textbf{Information Sufficiency Assessment}, evaluating whether the current information is adequate for producing a reliable answer; and (2) \textbf{Budget-Conditioned Strategy}, determining whether the next search is necessary and worthwhile given the observed budget state and current information. More details can be found in Appendix~\ref{app:prompt}.

\subsection{Curriculum Learning}
\label{sec:curriculum}

To internalize budget-aware search, we employ a two-phase curriculum~\cite{narvekar2020curriculum,bengio2009curriculum} via GRPO~\cite{shao2024deepseekmath} algorithm, progressing from budget annealing to adaptive sampling with a sliding window that focuses training on budget levels with lower accuracy while maintaining overall coverage.

\paragraph{Phase I: Annealing.}
We linearly decay the budget from $B_{max}$ to $1$ over this phase, allocating equal training steps to each budget level, with the scaffold (\cref{sec:scaffold}) active to guide exploration. The agent first learns \textit{how} to search under abundant budgets, then progressively learns \textit{when} to search as constraints tighten, while initializing the per-level accuracy statistics needed by Phase~II.

\paragraph{Phase II: Adaptive Budget Sampling.}
The scaffold is removed in this phase, and the agent only receives the total budget $B$ at the beginning of each episode.
After annealing, the budget is sampled from an adaptive distribution that focuses training on budget levels where the agent's performance remains weakest. We maintain a per-level sliding window of size $W$ tracking accuracy at each budget level $b \in \{1, \ldots, B_{max}\}$. At training step $t$:
\begin{equation}
    \bar{R}^{(t)}(b) = \frac{1}{|\mathcal{W}_b^{(t)}|} \sum_{\tau \in \mathcal{W}_b^{(t)}} \mathbb{I}_{ans}(\tau)
\end{equation}
where $\mathbb{I}_{ans}(\tau) \in \{0, 1\}$ denotes the answer accuracy of trajectory $\tau$ and $\mathcal{W}_b^{(t)}$ contains the most recent $W$ trajectories assigned budget $b$. The sampling distribution is:

\begin{align}
\label{eq:adaptive_sampling}
    P^{(t)}(B{=}b) &= (1{-}\lambda) \cdot \frac{\bar{R}_{max}^{(t)} - \bar{R}^{(t)}(b) + \epsilon}{\sum_{b'} \left(\bar{R}_{max}^{(t)} - \bar{R}^{(t)}(b') + \epsilon\right)} \nonumber \\
    &\quad + \lambda \cdot \frac{1}{B_{max}}
\end{align}
where $\bar{R}_{max}^{(t)} = \max_b \bar{R}^{(t)}(b)$, $\epsilon$ is a smoothing constant for numerical stability and $\lambda$ controls the minimum sampling probability assigned to each budget level. Intuitively, this distribution increases the sampling probability for budget levels where the agent lags furthest behind its best performance, while retaining minimum sampling exposure to all budget levels to ensure distributional robustness across all budget levels.

We optimize the policy via GRPO~\cite{shao2024deepseekmath}, with the full optimization objective detailed in Appendix~\ref{app:grpo}.

\subsection{Reward Design}
\label{sec:reward}

The composite reward combines four signals:
\begin{equation}
\label{eq:total_reward}
    R_{total} = \alpha \cdot R_{acc} + \beta \cdot R_{format} + \delta \cdot R_{length} + \gamma_q \cdot R_{tool}
\end{equation}
where $R_{acc}$ measures answer accuracy, $R_{format}$ enforces structural validity of the interaction trajectory, and $R_{length}$ applies a length penalty (details of these terms in Appendix~\ref{app:reward}). Below we focus on $R_{tool}$ and $\gamma_q$, which are central to incentivizing budget-aware search capability.

\paragraph{Tool Reward.}
$R_{tool}$ measures how efficiently the agent uses its budget, decomposing into an absolute signal and a relative signal:
\begin{equation}
    R_{tool} = R_{abs} \cdot R_{rel}
\end{equation}

The absolute signal $R_{abs}$ rewards correct answers proportionally to the budget saved:
\begin{equation}
    R_{abs} = \mathbb{I}_{ans} \cdot \frac{B_{total} - B_{used}}{B_{total}}
\end{equation}
This incentivizes the agent to answer correctly using the fewest budget possible.

The relative signal $R_{rel}$ compares against the most efficient correct trajectory in the same sampling group:
\begin{equation}
R_{rel} =
\mathbb{I}_{ans} \cdot \left(1 - \frac{B_{used} - B_{min}^+}{B_{max}^+ - B_{min}^+ + \xi}\right)
\end{equation}
where $B_{min}^+$ and $B_{max}^+$ are the minimum and maximum budget used among correct trajectories in the group, and $\xi$ is a stability constant.

\paragraph{Adaptive Efficiency Weight.}
We dynamically adjust $\gamma_q$ based on group accuracy to control the proportion of $R_{tool}$ in $R_{total}$:
\begin{equation}
\gamma_q = \gamma_{max} \cdot \frac{1}{G} \sum_{i=1}^{G} \mathbb{I}_{ans}^{(i)}
\end{equation}
where $G$ is the group size. For high-accuracy queries, $R_{acc}$ provides little differentiation and $R_{tool}$ dominates the advantage, driving the agent to optimize efficiency. But for low-accuracy queries, $\gamma_q$ is attenuated so that the advantage is primarily shaped by $R_{acc}$, directing the agent to prioritize answer accuracy over search efficiency.

\section{Experiments}

\subsection{Experimental Setup}

\begin{table*}[t]
    \footnotesize
    \centering
    \caption{Main results of EM scores across seven QA benchmarks using different backbone models. All methods are prompted with a search-call budget of $B=5$, which limits the maximum number of search actions. Results are means over three independent evaluations.}

    \label{tab:exp_main}
    \begin{tabularx}{\textwidth}{l RRR BBBB Y} 
        \toprule
        \rowcolor{white} 
        \textbf{Methods} 
        & \multicolumn{3}{c}{\cellcolor{bg_red}\textbf{General QA}} 
        & \multicolumn{4}{c}{\cellcolor{bg_blue}\textbf{Multi-Hop QA}} 
        & \\
        
        \cmidrule{2-4} \cmidrule{5-8}
        \rowcolor{white} 
         & \textbf{NQ$^\dagger$} & \textbf{TriviaQA$^*$} & \textbf{PopQA$^*$} 
         & \textbf{HotpotQA$^\dagger$} & \textbf{2wiki$^*$} & \textbf{Musique$^*$} & \textbf{Bamboogle$^*$} 
         & \makecell[c]{\textbf{Avg.}}\\
        \midrule
        
        \multicolumn{9}{l}{\textbf{Qwen2.5-7B-Instruct}} \\
        BATS & 0.223 & 0.428 & 0.128 & 0.284 & 0.205 & 0.051 & 0.272 & 0.227 \\
        Search-o1 & 0.266 & 0.579 & 0.286 & 0.240 & 0.229 & 0.076 & 0.184 & 0.266 \\
        Search-R1 & 0.388 & 0.620 & 0.501 & 0.372 & \underline{0.356} & 0.169 & \underline{0.372} & 0.397 \\
        
        ZeroSearch & 0.407 & 0.642 & \underline{0.522} & 0.328 & 0.315 & 0.178 & 0.319 & 0.387 \\
        StepSearch & \underline{0.415} & \underline{0.654} & 0.514 & \underline{0.384} & 0.342 & \underline{0.185} & 0.328 & \underline{0.403} \\
        \midrule  
        \rowcolor{row-highlight}\textbf{AnySearch} & \textbf{0.436} & \textbf{0.687} & \textbf{0.529} & \textbf{0.395} & \textbf{0.364} & \textbf{0.205} & \textbf{0.398} & \textbf{0.431} \\
        \midrule
        
        \multicolumn{9}{l}{\textbf{Llama-3.1-8B-Instruct}} \\
        BATS & 0.273 & 0.590 & 0.198 & 0.208 & 0.240 & 0.088 & 0.248 & 0.264 \\
        Search-o1 & 0.365 & 0.604 & 0.316 & 0.321 & 0.307 & 0.182 & 0.336 & 0.347 \\
        Search-R1 & 0.401 & 0.642 & 0.514 & \underline{0.364} & 0.349 & 0.188 & 0.365 & 0.403 \\
        ZeroSearch & \textbf{0.468} & \underline{0.664} & \underline{0.559} & 0.332 & \underline{0.368} & 0.184 & \underline{0.392} & \underline{0.424} \\
        StepSearch & \underline{0.422} & 0.635 & 0.531 & 0.350 & 0.361 & \underline{0.193} & 0.376 & 0.410 \\
        \midrule
        \rowcolor{row-highlight}\textbf{AnySearch} & \textbf{0.468} & \textbf{0.684} & \textbf{0.576} & \textbf{0.402} & \textbf{0.392} & \textbf{0.203} & \textbf{0.412} & \textbf{0.448} \\
        \midrule

        \multicolumn{9}{l}{\textbf{Qwen3-4B}} \\
        BATS & 0.278 & 0.384 & 0.112 & 0.165 & 0.210 & 0.045 & 0.232 & 0.204 \\
        Search-o1 & 0.318 & 0.586 & 0.231 & 0.268 & 0.258 & 0.064 & 0.280 & 0.286 \\
        Search-R1 & 0.375 & 0.604 & 0.448 & \underline{0.356} & \underline{0.353} & 0.132 & 0.344 & 0.373 \\

        ZeroSearch & 0.392 & 0.598 & 0.432 & 0.328 & 0.342 & 0.115 & 0.326 & 0.362 \\
        StepSearch & \underline{0.407} & \underline{0.616} & \underline{0.465} & 0.345 & 0.349 & \underline{0.148} & \textbf{0.352} & \underline{0.383} \\
        \midrule
        \rowcolor{row-highlight}\textbf{AnySearch} & \textbf{0.429} & \textbf{0.648} & \textbf{0.496} & \textbf{0.384} & \textbf{0.375} & \textbf{0.164} & \textbf{0.352} & \textbf{0.407} \\
        \midrule
        
        \rowcolor{white}\multicolumn{9}{l}{\small \textbf{Bold} indicates the best, \underline{underline} indicates the second-best performance within each backbone group.} \\
        \rowcolor{white}\multicolumn{9}{l}{\small $^\dagger$ indicates in-domain evaluation; $^*$ indicates out-of-domain evaluation.} \\
    \end{tabularx}
\end{table*}
\begin{table}[t]
\centering
\caption{Comparison of efficiency on HotpotQA under different settings measured by Accuracy and TP.}
\label{tab:effiency}
\resizebox{\linewidth}{!}{
\begin{tabular}{cccc}
\toprule
\rowcolor{gray!15}
\textbf{Settings} & \textbf{Methods} & \textbf{Accuracy$\uparrow$} & \textbf{TP$\uparrow$} \\
\midrule
\multirow{3}{*}{\makecell{Qwen2.5-7B\\B = 6}} & \textbf{AnySearch}         & \textbf{0.398}  & \textbf{0.354}  \\
                   & Search-R1    & 0.375  & 0.212  \\
                   & StepSearch   & 0.388  & 0.276  \\

\midrule
\multirow{3}{*}{\makecell{Llama3.1-8B\\B = 5}} & \textbf{AnySearch}         & \textbf{0.402}  & \textbf{0.378}  \\
                   & Search-R1    & 0.364  & 0.236  \\
                   & StepSearch   & 0.350  & 0.297  \\
\midrule
\multirow{3}{*}{\makecell{Qwen3-4B\\B = 4}} & \textbf{AnySearch}         & \textbf{0.378}  & \textbf{0.309}  \\
                   & Search-R1    & 0.351  & 0.235  \\
                   & StepSearch   & 0.339  & 0.198  \\
\bottomrule
\end{tabular}
}
\end{table}

\begin{figure}
\centering
\includegraphics[width=\linewidth]{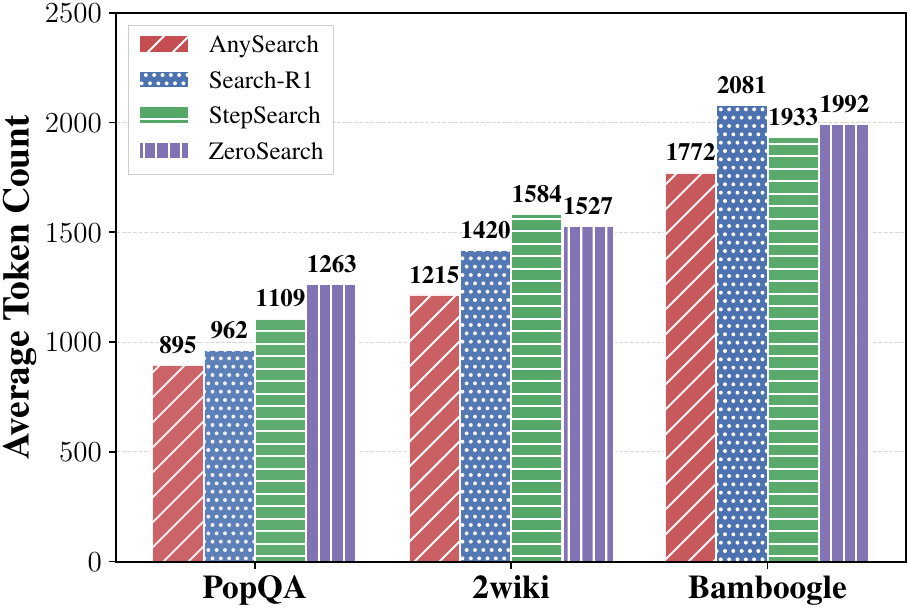}
\caption{Average token count across different benchmarks on Llama-3.1-8B-Instruct. Here, the token count is the sum of output token and retrieved token.}
\label{fig:token}
\end{figure}
\begin{table}[t]
\centering
\caption{Ablation study of the tool reward on 2wiki with Qwen3-4B, covering the adaptive efficiency weight $\gamma_q$ and the reward components $R_{abs}$ and $R_{rel}$.}
\label{tab:tool-reward}
\begin{tabular}{lcc}
\toprule
\rowcolor{gray!15}
\textbf{Settings} & \textbf{Accuracy$\uparrow$} & \textbf{TP$\uparrow$} \\
\midrule
\textbf{AnySearch (Adaptive $\gamma_q$)} & \textbf{0.375} & \textbf{0.342} \\
Fixed $\gamma_q = \gamma_{max}$  & 0.354 & 0.328 \\
Fixed $\gamma_q = 0.5 \cdot \gamma_{max}$ & 0.366 & 0.334 \\
\textit{w/o} $R_{tool}$ (Fixed $\gamma_q = 0$) & 0.348 & 0.226 \\
\textit{w/o} $R_{abs}$ & 0.352 & 0.272 \\
\textit{w/o} $R_{rel}$ & 0.359 & 0.295 \\
\bottomrule
\end{tabular}
\end{table}

\begin{figure*}[t]
\centering
\makebox[\textwidth][c]{%

\begin{subfigure}[b]{0.31\textwidth}
\centering
\includegraphics[width=\linewidth]{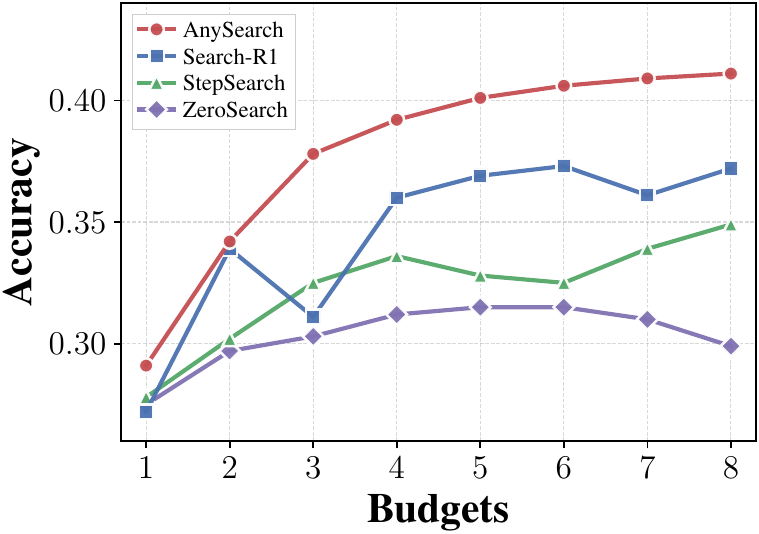}
\caption{Qwen2.5-7B on Bamboogle}
\end{subfigure}%
\hfill

\begin{subfigure}[b]{0.31\textwidth}
\centering
\includegraphics[width=\linewidth]{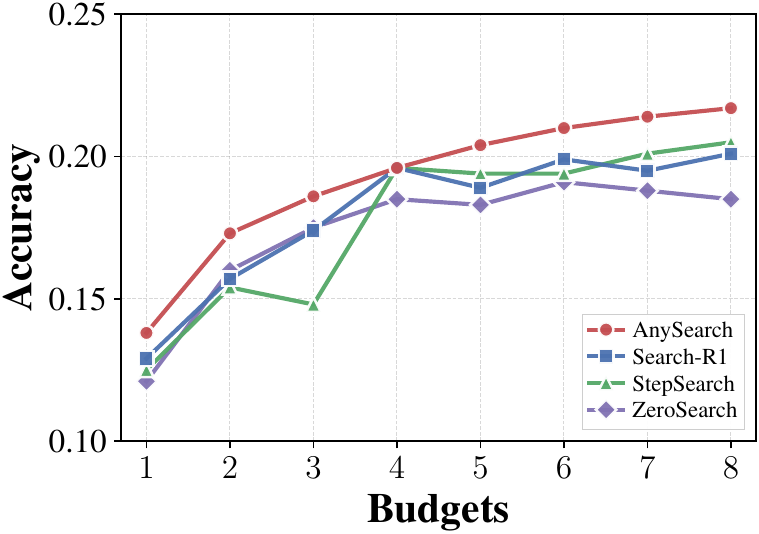}
\caption{Llama-3.1-8B on Musique}
\end{subfigure}%
\hfill

\begin{subfigure}[b]{0.31\textwidth}
\centering
\includegraphics[width=\linewidth]{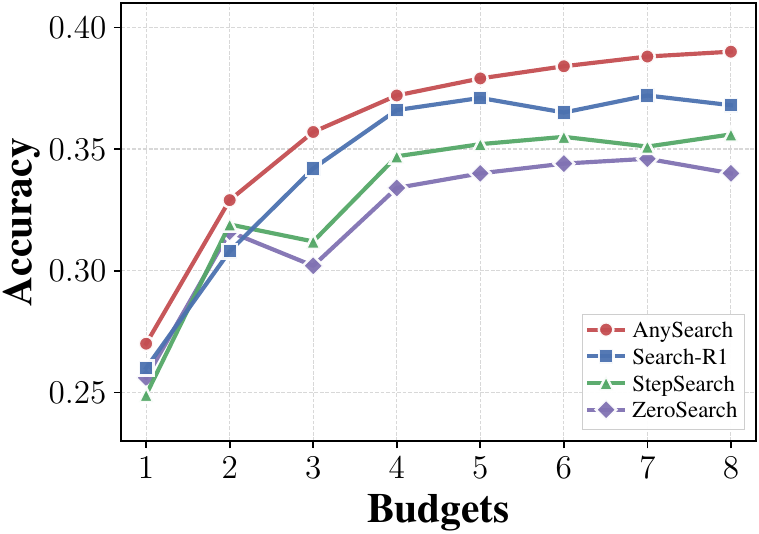}
\caption{Qwen3-4B on 2wiki}
\end{subfigure}%
}
\caption{Performance of RL-based methods across varying budgets. We report the accuracy curves for (a) Qwen2.5-7B-Instruct on Bamboogle, (b) Llama-3.1-8B-Instruct on Musique, and (c) Qwen3-4B on 2wiki. For tool-free baselines such as ZeroSearch, we define the budget as the max turn, set to $B+1$.}
\label{fig:anybudget}
\end{figure*}
\begin{figure*}[t]
\centering
\begin{subfigure}{0.31\linewidth}
\centering
\includegraphics[width=\linewidth]{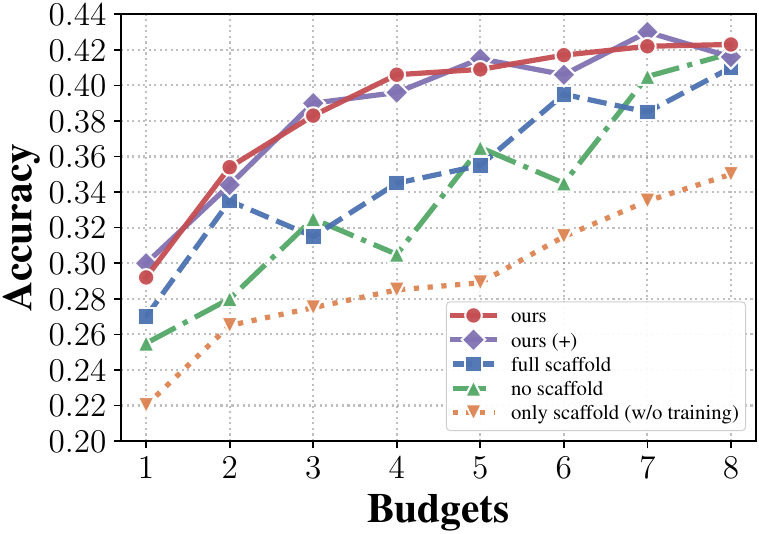}
\caption{Scaffold ablation}
\label{subfig:scaffold}
\end{subfigure}
\hfill
\begin{subfigure}{0.31\linewidth}
\centering
\includegraphics[width=\linewidth]{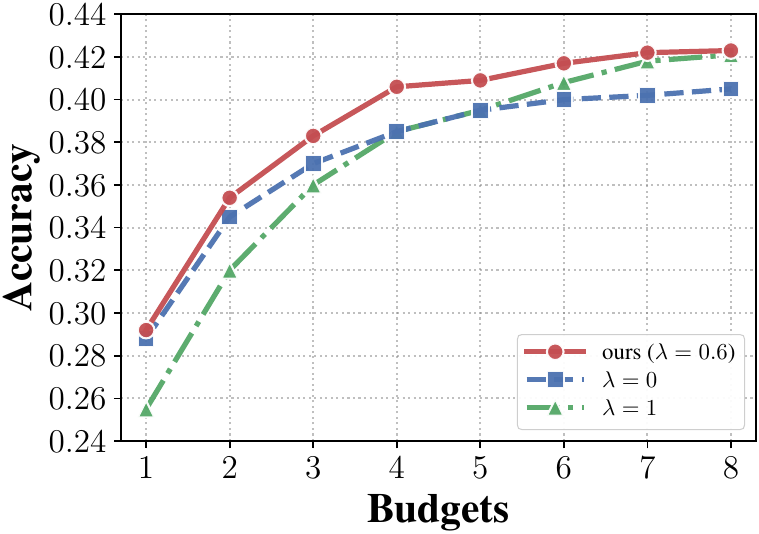}
\caption{Sampling strategy ablation}
\label{subfig:lambda}
\end{subfigure}
\hfill
\begin{subfigure}{0.31\linewidth}
\centering
\includegraphics[width=\linewidth]{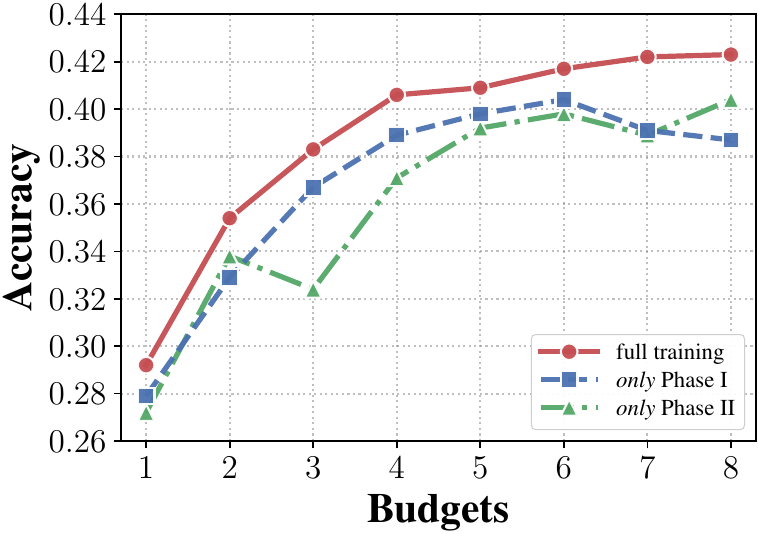}
\caption{Curriculum phase ablation}
\label{subfig:phase}
\end{subfigure}
\caption{Ablation studies on HotpotQA with Qwen2.5-7B-Instruct across varying budgets. (a)~Scaffold ablation. Full Scaffold uses scaffold in both Phase~I and Phase~II but removes it at inference; No Scaffold never uses scaffold in training or inference; Ours uses scaffold in Phase~I, removes it in Phase~II and inference; Ours (+Scaffold) applies our full training and retains the scaffold at inference; Only Scaffold uses scaffold at inference without any RL training. (b)~Budget sampling strategy in Phase~II, comparing $\lambda{=}0$, $\lambda{=}1$, and Ours ($\lambda{=}0.6$). (c)~Curriculum phase ablation, comparing training with only Phase~I, only Phase~II, and full training, all trained for 500 steps.}
\label{fig:ablation}
\end{figure*}

\begin{figure*}[t]
\centering
\makebox[\textwidth][c]{%

\begin{subfigure}[b]{0.31\textwidth}
\centering
\includegraphics[width=\linewidth]{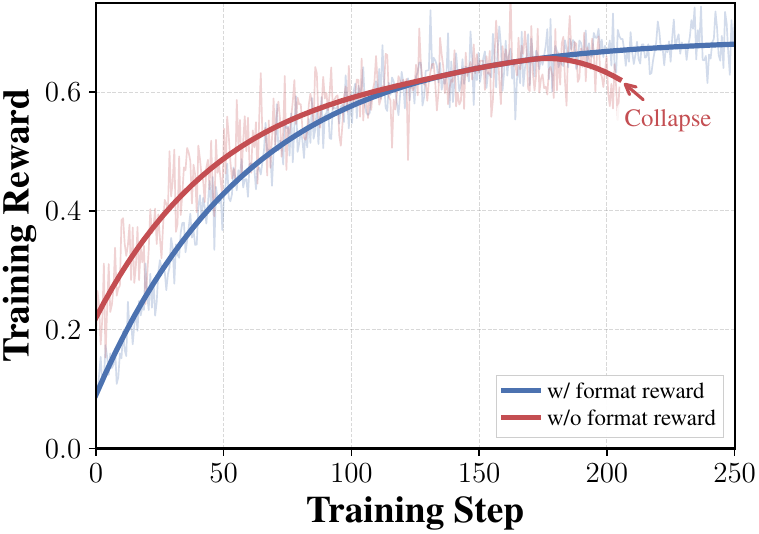}
\caption{Training Reward Curve}
\end{subfigure}%
\hfill

\begin{subfigure}[b]{0.31\textwidth}
\centering
\includegraphics[width=\linewidth]{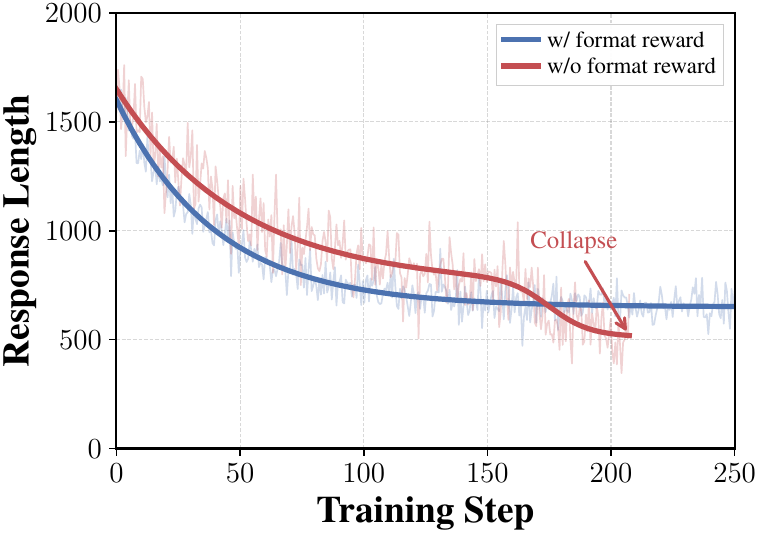}
\caption{Response Length Curve}
\end{subfigure}%
\hfill

\begin{subfigure}[b]{0.31\textwidth}
\centering
\includegraphics[width=\linewidth]{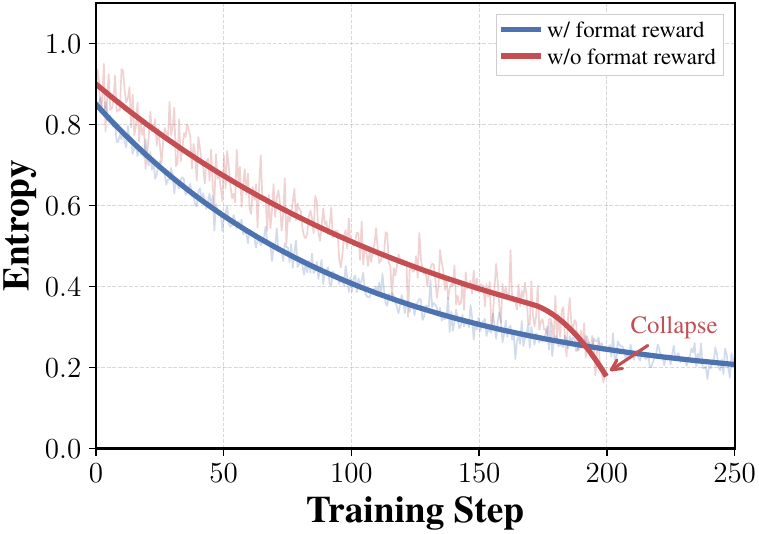}
\caption{Entropy Curve}
\end{subfigure}%
}

\caption{Training dynamics on Qwen3-4B showing the role of format rewards in training stability. (a) Without $R_{format}$, training reward rises initially but reverses around training step 175, while the full setting remains stable. (b) The no-format variant undergoes a sharp response-length reduction around the same point, indicating degeneration into trivial short outputs. (c) Its policy entropy also collapses abruptly, whereas the full setting decays smoothly and retains exploration.}
\label{fig:dynamic}
\end{figure*}

\paragraph{Training Details.}
We use Qwen2.5-7B-Instruct~\cite{qwen2}, Llama-3.1-8B-Instruct~\cite{grattafiori2024llama}, and Qwen3-4B~\cite{yang2025qwen3} as backbone models. For RL training, we adopt GRPO~\cite{shao2024deepseekmath} implemented with the \textit{Slime} framework~\cite{slime_github}. Following~\cite{jin2025search}, we use the 2018 Wikipedia dump~\cite{karpukhin2020dense} as the retrieval corpus and employ E5~\cite{wang2022text} as the dense retriever with top-$k=3$ documents per query. Training budget is set to $B_{\max}=5$. We train on NQ~\cite{kwiatkowski2019natural} and HotpotQA~\cite{yang2018hotpotqa} to cover both general and multi-hop QA. Full hyperparameter details are in Appendix~\ref{app:implementation}.

\paragraph{Evaluation Details.}
We evaluate on seven benchmarks: three general QA (NQ, TriviaQA~\cite{joshi2017triviaqa}, PopQA~\cite{mallen2023not}) and four multi-hop QA (HotpotQA, 2WikiMultiHopQA~\cite{ho2020constructing}, MuSiQue~\cite{trivedi2022musique}, Bamboogle~\cite{press2023measuring}). We report \textbf{Exact Match (EM)} as the primary metric for fair comparison. To measure retrieval efficiency,  we also report \textbf{Tool Productivity (TP)} following~\cite{wang2025acting}, defined as the number of correctly answered questions per unit of external search. Formally, TP is computed as:
\begin{equation}
    \text{TP} = \frac{\sum_{i=1}^{N} \mathbb{I}\{ans_i = y_i^*\}}{\sum_{i=1}^{N} c_i}
\end{equation}
where $c_i$ is the search count for instance $i$.

\paragraph{Baselines.}
We compare against prompt-based budget-aware methods (BATS~\cite{liu2025budget}), RAG-based methods (Search-o1~\cite{li2025search}), and RL-based methods (Search-R1~\cite{jin2025search}, ZeroSearch~\cite{sun2025zerosearch}, StepSearch~\cite{wang2025stepsearch}). All methods are prompted with the search budget that limits the maximum number of search actions. 
More details are provided in Appendix~\ref{app:baselines}.

\subsection{Main Results}

\textbf{Our method delivers consistent improvements over all baselines on both general and multi-hop QA.} As shown in \cref{tab:exp_main}, using Qwen2.5-7B-Instruct as the backbone, our method achieves an average EM of 0.431, outperforming the strongest RL baseline StepSearch at 0.403. BATS and Search-o1, which lack RL-based policy optimization, fall considerably behind at 0.227 and 0.266 respectively, confirming that budget awareness alone without trained allocation is insufficient. Our advantage holds across both in-domain and out-of-domain datasets, and similar trends persist on Llama-3.1-8B-Instruct and Qwen3-4B, demonstrating consistent effectiveness across different model families and sizes. 

\textbf{Our method consistently outperforms baselines across varying budget scales, generalizing effectively to constraints unseen during training.} According to \cref{fig:anybudget}, our method maintains a consistently dominant and monotonically increasing performance curve across the entire spectrum, including the unseen budgets of 6 to 8 that exceed our maximum training budget of 5. In contrast, the baselines exhibit noticeable fluctuations with occasional mid-range drops and tend to plateau at higher budgets. Specifically, on Bamboogle using Qwen2.5-7B-Instruct, our method achieves an accuracy of about 0.40 at a budget of 5 and further improves to about 0.42 at a budget of 8, whereas the strongest baseline Search-R1 plateaus around 0.37. This sustained superiority at unseen budgets suggests that the agent has internalized budget-proportional allocation rather than memorizing a fixed search strategy.

\textbf{Our method achieves superior accuracy with fewer search calls, pushing the cost-performance Pareto frontier beyond all baselines.} As shown in \cref{fig:anybudget}, on Bamboogle using Qwen2.5-7B-Instruct at the same budget of 3, our method achieves a score of 0.38, which is higher than Search-R1 at 0.34. Conversely, to reach the accuracy of about 0.35, the strongest baseline requires a budget of 4, whereas our method meets the same threshold with a budget of 3. These results suggest that budget-aware training produces more effective allocation decisions, yielding stable performance improvements under any constraint.

\subsection{Analysis of Efficiency}

\cref{tab:effiency} shows that our method achieves the highest Tool Productivity across all backbone models and budget levels, meaning it answers more questions correctly per external search invoked. Furthermore, as retrieval documents dominate total token consumption (see \cref{tab:token-ratio} in Appendix), reducing search calls directly lowers overall inference cost. As shown in \cref{fig:token}, our method achieves the lowest total token count across all datasets, indicating that budget-aware training enables more effective utilization of each search call and avoids redundant retrieval. Full token consumption results are provided in Appendix~\ref{app:token}, and full efficiency comparisons across all multi-hop benchmarks are in Appendix~\ref{app:efficiency}. The training-cost comparison in Appendix~\ref{app:training_cost} further shows that our two-phase curriculum has a single-run cost comparable to fixed-budget baselines while amortizing one policy across the tested budget range.

\subsection{Ablation Studies}

\textbf{Tool reward.}
\cref{tab:tool-reward} ablates the tool reward along two dimensions. For the adaptive efficiency weight, fixing $\gamma_q = \gamma_{max}$ causes accuracy to drop while TP increases, as the agent over-prioritizes efficiency on difficult queries where it should focus on answering correctly. Fixing $\gamma_q = 0.5 \cdot \gamma_{max}$ partially alleviates this but remains suboptimal. Removing $R_{tool}$ entirely ($\gamma_q = 0$) leads to the sharpest accuracy and TP decline, confirming that explicit efficiency feedback is essential. The adaptive design resolves this tension by naturally attenuating $\gamma_q$ on low-accuracy queries so that the agent prioritizes correctness, while on high-accuracy queries the efficiency signal dominates and drives the agent to reduce redundant searches. For the reward components, removing $R_{abs}$ eliminates the absolute incentive to save budget, while removing $R_{rel}$ removes the relative pressure to approach the most efficient correct trajectory within each group. Both provide complementary efficiency signals and are jointly necessary.

\textbf{Early-stopping analysis.} The efficiency reward is correctness-gated because both components of $R_{tool}$ are multiplied by $\mathbb{I}_{ans}$, so an incorrect early answer receives no efficiency reward. Moreover, $\gamma_q$ decreases on low-accuracy groups, reducing efficiency pressure on hard queries. We compare AnySearch $\pi$ with a $\gamma_q=0$ control $\pi_0$ that removes the efficiency reward while keeping the framework and training configuration fixed. On 2Wiki with Qwen3-4B under a budget of five searches, AnySearch is incorrect while the control is correct on 77 examples. Only 41 cases, or 0.33\% of the test set, terminate with unused budget. In contrast, AnySearch correctly answers 417 examples missed by the control, a gain of 3.32\%. This yields a net gain of 340 examples, or 2.70\%, as shown in \cref{tab:early_stop}. Thus, early stopping is rare and is substantially outweighed by the accuracy gains from better allocation.

\begin{table}[t]
\centering
\small
\caption{Counterfactual analysis of early stopping on 2Wiki with Qwen3-4B under a budget of five searches. $\pi$ is AnySearch and $\pi_0$ is the $\gamma_q=0$ control without efficiency reward.}
\label{tab:early_stop}
\begin{tabular}{lcc}
\toprule
\textbf{Outcome} & \textbf{Count} & \textbf{Rate} \\
\midrule
$\pi$ incorrect, $\pi_0$ correct & 77 & 0.61\% \\
Cases with unused budget ($B_{used}^{\pi}<B$) & 41 & 0.33\% \\
$\pi$ correct, $\pi_0$ incorrect & 417 & 3.32\% \\
Net accuracy gain & 340 & 2.70\% \\
\bottomrule
\end{tabular}
\end{table}

\textbf{Budget-aware search internalization.}
\cref{subfig:scaffold} validates the internalization hypothesis. Training with the scaffold throughout but removing it at inference (Full Scaffold) leads to noticeable performance degradation, particularly at low budgets where budget-aware allocation matters most. This indicates that the agent becomes reliant on explicit budget state signals without learning to allocate autonomously, producing a train-inference gap. Training without any scaffold (No Scaffold) results in the weakest overall performance, as the agent lacks sufficient guidance during early exploration to discover budget-aware decision patterns. Our strategy achieves the highest accuracy across all budget levels with a smooth, monotonically increasing curve. The advantage is most pronounced under tight budgets, confirming that the scaffold successfully guides the agent toward budget-aware behaviors in Phase~I, and these behaviors are retained after scaffold removal in Phase~II. Moreover, Ours (+) shows that the trained policy no longer benefits from the scaffold at inference, while Only Scaffold shows that the scaffold without RL training is ineffective, jointly confirming successful internalization.

\textbf{Adaptive budget sampling.}
\cref{subfig:lambda} compares budget sampling strategies in Phase~II. Uniform sampling ($\lambda{=}1$) distributes training equally across all budget levels, resulting in weaker performance at low budgets where the task is inherently more difficult and requires more training exposure. Adaptive sampling without uniform smoothing ($\lambda{=}0$) focuses entirely on budget levels with low accuracy, which improves low-budget performance but causes degradation at high budgets due to insufficient coverage of well-learned levels. Our full adaptive distribution ($\lambda{=}0.6$) achieves the best performance across all budget levels, balancing focused training on weak levels with maintained coverage of strong ones. The resulting sampling distribution is visualized in Appendix~\ref{app:sampling}.

\textbf{Curriculum phases.}
\cref{subfig:phase} ablates the two training phases. Training with only Phase~I provides budget annealing but lacks adaptive sampling, while training with only Phase~II applies adaptive sampling without warm-up. Both configurations underperform and fail to improve steadily with budget. Full two-phase training achieves the highest performance and the most stable budget-scaling curve, validating that Phase~I teaches budget-aware search under scaffold guidance while Phase~II removes the scaffold and consolidates robust performance across all budgets.

\subsection{Discussion}
\textbf{Training stability.}
\label{sec:exp_discussion}
\cref{fig:dynamic} highlights the stabilizing role of $R_{format}$. Removing this component causes the reward to deteriorate around training step 175, together with sharp decreases in response length and policy entropy. These trends indicate a collapse toward overly short outputs with limited exploration. In contrast, the full reward formulation maintains stable learning and well-formed trajectories throughout training. The same pattern is observed for the other backbone models in Appendix~\ref{app:training_dynamics}.

\textbf{Case Study.} Appendix~\ref{app:cases} presents qualitative examples showing that the trained agent adjusts search depth adaptively to budget availability, confirming internalization of budget-aware search.

\section{Conclusion}
We presented AnySearch, a framework that internalizes budget-aware search into a single policy via a progressively removed training scaffold and a two-phase curriculum RL with adaptive budget sampling. A composite reward with adaptive efficiency weighting jointly optimizes accuracy and search efficiency. Experiments across seven general and multi-hop QA benchmarks and three backbone models show consistent improvements over baselines at all budget levels, robust generalization to unseen budgets, highest tool productivity, and lowest token consumption. Ablations confirm that the scaffold is successfully internalized and each component contributes to the final performance.

\section*{Limitations}
Although AnySearch internalizes budget-aware search into a single policy and adapts across budgets without retraining, several limitations remain. First, we model the budget as a discrete count of external search calls. In real deployment the true cost is multidimensional, spanning latency, monetary expense, and system load, so a continuous and multi-objective cost formulation is a natural extension of our framework. Second, the quality of budget allocation is still bounded by the backbone model. AnySearch decides when and what to search more efficiently, but it cannot recover an answer that lies neither in the model's parametric knowledge nor in the retrieval corpus. Finally, our training and evaluation use a static 2018 Wikipedia dump, following the standard protocol of prior RL-search agents. This keeps the retrieval environment controlled and reproducible, but it does not test temporal adaptation, and extending AnySearch to a continuously updated open-web setting is left to future work.

\section*{Ethics Statement}
Our research adheres to strict ethical guidelines. We verified the licenses of all software, models, and datasets used in this study to ensure full compliance with their terms. Since this work focuses on training search agents via reinforcement learning using existing public QA datasets and a publicly available Wikipedia corpus, no human annotation or primary data collection was involved. No privacy concerns or personally identifiable information have been identified in the training or evaluation data. Additionally, we have conducted a thorough assessment of the project and do not anticipate any further risks.

\section*{Acknowledgments}
We gratefully acknowledge the insightful discussions, constructive feedback, and support received throughout this work.

\bibliography{acl}

\appendix
\clearpage
\section{Experiment Details}
\label{app:exp_details}

In this section, we provide a comprehensive breakdown of our experimental setup, ensuring reproducibility and clarifying the configurations used for all baselines and our proposed method.

\subsection{Implementation Details and Experimental Configuration}
\label{app:implementation}

Our implementation is built upon the \textbf{Slime} framework~\cite{slime_github}, leveraging \textbf{GRPO}~\cite{shao2024deepseekmath} within an asynchronous distributed training architecture.

\begin{table*}[t]
\centering
\caption{Key hyperparameter configuration for our experiments.}
\label{tab:comprehensive_hyperparams}
\renewcommand{\arraystretch}{1.15}
\begin{tabular}{ll|c}
\toprule
\rowcolor{gray!15} \textbf{Category} & \textbf{Parameter} & \textbf{Value} \\
\midrule
\textbf{Optimization} & Learning Rate & $1.0 \times 10^{-6}$ \\
& Optimizer Betas & $\beta_1=0.9, \beta_2=0.98$ \\
& Weight Decay & 0.01 \\
\midrule
\textbf{Generation} & Group Size ($G$) & 5 \\
& Global batch size & 512 \\
& KL Penalty Coefficient ($\beta_{KL}$) & 0.001 \\
& Policy Clip Ratio  & 0.2 (low) / 0.28 (high) \\
& Advantage Stability ($\epsilon_0$) & 1e-6 \\
& Max Response Length & 4096 \\
& Sampling Temperature & 1.0 \\
& Top-p & 1.0 \\

\midrule
\textbf{Rewards} & Accuracy Reward ($\alpha$) & 0.5 \\
& Format Reward ($\beta$) & 0.15 \\
& Length Reward ($\delta$) & 0.05 \\
& Tool Reward ($\gamma_{max}$) & 0.3 \\
& Stability Constant ($\xi$) & 1e-6 \\
\midrule
\textbf{Curriculum} & Phase I (Warm-up) & 100 Steps \\
& Phase II (Adaptive Sampling) & 400 Steps \\
& $B_{max}$ & 5 \\
& Sliding Window Size ($W$) & 20 \\
& Uniform Smoothing ($\lambda$) & 0.6 \\
& Smoothing Constant ($\epsilon$) & 1e-6 \\
\bottomrule
\end{tabular}
\end{table*}

\begin{table*}[t]
\centering
\caption{Statistics of the datasets used in our experiments.}
\label{tab:dataset_stat}
\begin{tabular}{lccc}
\toprule
\rowcolor{gray!15} \textbf{Dataset} & \textbf{Type} & \textbf{Size} & \textbf{Source} \\
\midrule
\multicolumn{4}{l}{\textit{Training Sets}} \\
Natural Questions (NQ) & General QA & 79k & \citet{kwiatkowski2019natural} \\
HotpotQA & Multi-hop QA & 90k & \citet{yang2018hotpotqa} \\
\midrule
\multicolumn{4}{l}{\textit{Evaluation Sets}} \\
Natural Questions (NQ) & General QA & 3.6k & \citet{kwiatkowski2019natural} \\
TriviaQA & General QA & 11k & \citet{joshi2017triviaqa} \\
PopQA & General QA & 14k & \citet{mallen2023not} \\
HotpotQA & Multi-hop QA & 7.4k & \citet{yang2018hotpotqa} \\
2WikiMultiHopQA & Multi-hop QA & 12k & \citet{ho2020constructing} \\
Musique & Multi-hop QA & 2.4k & \citet{trivedi2022musique} \\
Bamboogle & Multi-hop QA & 125 & \citet{press2023measuring} \\
\bottomrule
\end{tabular}
\end{table*}

\subsubsection{Asynchronous Distributed Training Framework}
We use the asynchronous architecture of the Slime framework orchestrated via Ray. By decoupling the training actor from rollout workers, GPU idle time is eliminated. Experiments run on a single node with eight NVIDIA H800 (80GB) GPUs: GPUs 0--3 handle training (TP=2), and GPUs 4--7 handle rollout and retrieval. We use gradient checkpointing, sequence parallelism, and SGLang with a static memory fraction of 0.6.

\subsubsection{Retrieval Service Infrastructure}
We use E5-base-v2~\cite{wang2022text} as the dense retriever over the 2018 Wikipedia dump~\cite{karpukhin2020dense} ($\sim$18M passages), indexed via FAISS Flat for exact nearest-neighbor search. The service runs on port 8000 and returns top-$k=3$ documents per query formatted into \texttt{<information>} tags.

\subsubsection{Curriculum Learning Strategy}
We adopt a two-phase curriculum that progressively internalizes budget-aware search.

In \textbf{Phase I (Warm-up)} (first 100 steps, $\sim$20\% of training), the search budget is linearly decayed from $B_{max}=5$ to $1$, with equal training steps (20 steps per level) allocated to each budget level. The full scaffold (budget state injection and structured reasoning prompts) is active throughout this phase, guiding the agent to learn search behaviors across all budget levels.

In \textbf{Phase II (Adaptive Budget Sampling)} (remaining 400 steps), the scaffold is entirely removed and the agent only receives the total budget $B$ in natural language at the beginning of each episode. Each budget level independently tracks its accuracy over the most recent $W$ assigned episodes. The sampling probability is then set proportionally to the gap between each level's accuracy and the best-performing level, smoothed by $\epsilon$, and mixed with a uniform distribution weighted by $\lambda$ (see Eq.~\ref{eq:adaptive_sampling} for the full formulation).

\subsubsection{Training Cost Analysis}
\label{app:training_cost}

\begin{table*}[!t]
\centering
\small
\caption{Training cost comparison on one 8$\times$H800 node. All methods use 500 training steps. Fixed-budget methods require $K$ separately trained policies to serve $K$ budget levels, whereas AnySearch uses one policy across the tested budget range.}
\label{tab:training_cost}
\begin{tabular}{lccc}
\toprule
\textbf{Method} & \textbf{Total GPU-hours per Run} & \textbf{Policies for $K$ budgets} & \textbf{Supports varying budgets} \\
\midrule
Search-R1 & 320 & $K$ & No \\
StepSearch & 336 & $K$ & No \\
ZeroSearch & 304 & $K$ & No \\
\rowcolor{row-highlight}\textbf{AnySearch} & \textbf{328} & \textbf{1} & \textbf{Yes} \\
\bottomrule
\end{tabular}
\end{table*}

AnySearch uses 500 training steps, the same number as the fixed-budget baselines under the same per-step configuration. Its single-run cost is therefore comparable rather than additive. When deployment requires $K$ budget levels, fixed-budget baselines require $K$ separately trained policies, whereas one AnySearch policy serves the tested budget range, including the unseen budgets evaluated in \cref{fig:anybudget}. The resulting cost is amortized across budget levels rather than incurred once per level.

\subsubsection{GRPO Optimization Objective}
\label{app:grpo}

We optimize the policy $\pi_\theta$ via GRPO. For each query $q$ with assigned budget $B$, we sample $G$ trajectories $\{\tau_i\}_{i=1}^G$ from $\pi_{\theta_{\text{old}}}$ and compute advantages by standardizing the composite reward against group statistics:
\begin{equation}
    \hat{A}_i = \frac{R(\tau_i) - \text{mean}(\{R(\tau_j)\}_{j=1}^G)}{\text{std}(\{R(\tau_j)\}_{j=1}^G) + \epsilon_0}
\end{equation}
The policy is updated to maximize:
\begin{multline}
    \mathcal{J}(\theta) = \mathbb{E}_{q, B} \Bigg[ \frac{1}{G} \sum_{i=1}^G \frac{1}{|\tau_i|} \sum_{t=1}^{|\tau_i|} \Big( \min \big( r_{i,t} \hat{A}_i, \\
    \text{clip}(r_{i,t}, 1{-}\epsilon, 1{+}\epsilon) \hat{A}_i \big) - \beta_{KL} D_{KL}^{i,t} \Big) \Bigg]
\end{multline}
where $r_{i,t} = \pi_\theta(\tau_{i,t}|q, \tau_{i,<t}) / \pi_{\theta_{\text{old}}}(\tau_{i,t}|q, \tau_{i,<t})$ is the token-level importance ratio, $D_{KL}^{i,t} = \log \frac{\pi_\theta(\tau_{i,t}|q, \tau_{i,<t})}{\pi_{\text{ref}}(\tau_{i,t}|q, \tau_{i,<t})}$ is the per-token KL penalty, $\epsilon_0$ is a small constant for numerical stability, $\epsilon$ clips the ratio to stabilize updates, and $\beta_{KL}$ controls the strength of regularization toward the reference policy $\pi_{\text{ref}}$.

\subsubsection{Reward Function Configuration}
\label{app:reward}
The total reward is $R_{total} = \alpha \cdot R_{acc} + \beta \cdot R_{format} + \delta \cdot R_{length} + \gamma_q \cdot R_{tool}$, with the following components:
\begin{itemize}
     \item \textbf{Accuracy Reward ($\alpha=0.5$):} We extract the answer from the \texttt{<answer>} tag and compute a binary exact match against the ground truth $y^*$: $R_{acc} = \text{EM}(ans, y^*) \in \{0, 1\}$, returning 1 for a correct match and 0 otherwise.
    \item \textbf{Format Reward ($\beta=0.15$):} Defined as the product of three binary indicators $R_{format} = \prod_{i=1}^{3} \mathbb{I}_i$, yielding 1 only when all three constraints are satisfied and 0 otherwise:
    \begin{enumerate}
        \item[(i)] Proper pairing of all special tags. In Phase~I: \texttt{<budget>}, \texttt{<think>}, \texttt{<search>}, \texttt{<information>}, \texttt{<answer>}. In Phase~II: \texttt{<think>}, \texttt{<search>}, \texttt{<information>}, \texttt{<answer>} (no \texttt{<budget>} tag).
        \item[(ii)] Correct sequential ordering. In Phase~I: each round begins with \texttt{<budget>}, \texttt{<search>} is preceded by \texttt{<think>}, and the trajectory ends with \texttt{<answer>}. In Phase~II: \texttt{<search>} is preceded by \texttt{<think>}, and the trajectory ends with \texttt{<answer>}.
        \item[(iii)] No token generation outside valid tags.
    \end{enumerate}
    \item \textbf{Length Reward ($\delta=0.05$):} Following DAPO~\cite{yu2025dapo}, we apply a piecewise linear length penalty to discourage excessively long generations:
    \begin{equation}
    R_{length}(y) =
    \begin{cases}
    0 & |y| \le L_{lim} \\
    \frac{L_{lim} - |y|}{L_{tol}} & L_{lim} < |y| \le L_{lim} + L_{tol} \\
    -1 & |y| > L_{lim} + L_{tol}
    \end{cases}
    \end{equation}
    where $|y|$ denotes the generation length, $L_{limit}$ is the penalty threshold, and $L_{tol}$ is the tolerance window within which the penalty increases linearly from 0 to $-1$. In our experiments, we set $L_{limit} = 2048$ and $L_{tol} = 1024$.
    \item \textbf{Tool Reward ($\gamma_{max}=0.3$):} Decomposes into $R_{tool} = R_{abs} \cdot R_{rel}$. The absolute signal $R_{abs} = \mathbb{I}_{ans} \cdot (B_{total} - B_{used}) / B_{total}$ rewards correct answers proportionally to budget saved. The relative signal $R_{rel} = \mathbb{I}_{ans} \cdot (1 - (B_{used} - B_{min}^+) / (B_{max}^+ - B_{min}^+ + \xi))$ compares against the most and least efficient correct trajectories within the GRPO group. The weight $\gamma_q = \gamma_{max} \cdot \frac{1}{G}\sum_i \mathbb{I}_{ans}^{(i)}$ is scaled by group accuracy so that efficiency pressure is strong only when the agent already answers well.
\end{itemize}

\subsection{Full Efficiency Results}
\label{app:efficiency}

\cref{tab:efficiency_full} extends the efficiency comparison in the main text to all four multi-hop QA benchmarks across three backbone models and three budget levels ($B=4,5,6$). AnySearch consistently achieves the highest accuracy and tool productivity across all configurations, confirming that the efficiency advantage generalizes broadly beyond a single dataset or model.

\subsection{Token Consumption Analysis}
\label{app:token}

\begin{figure*}[t]
\centering
\includegraphics[width=\textwidth]{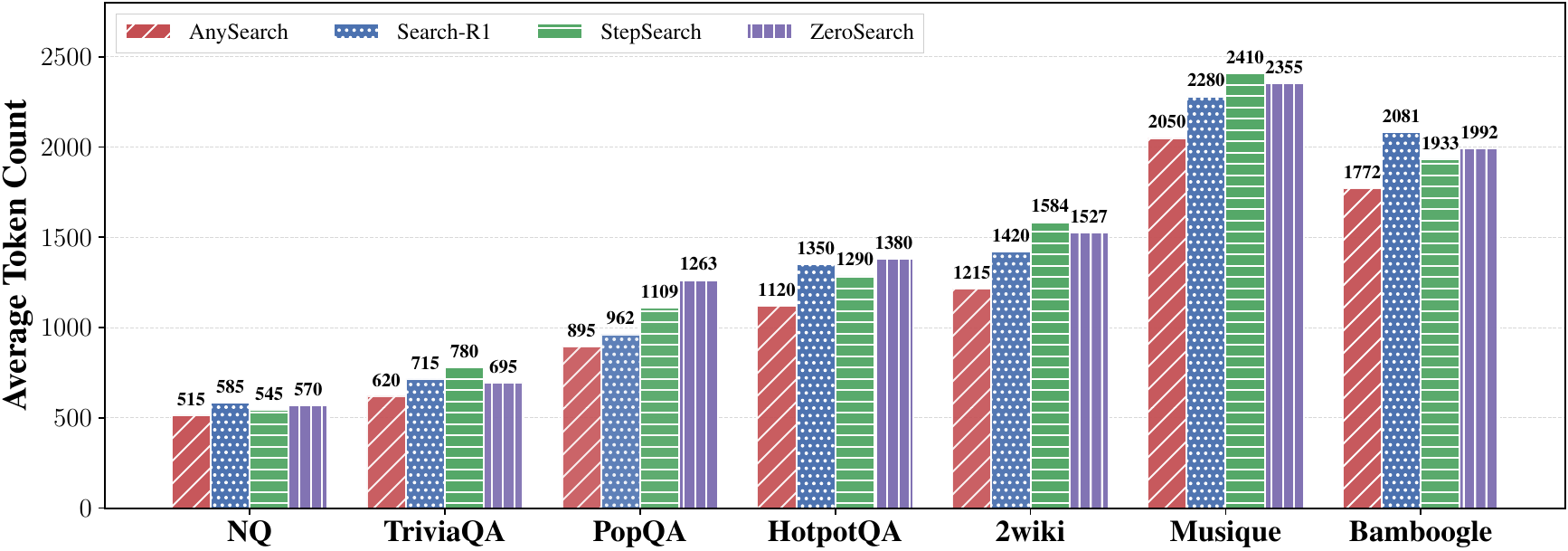}
\caption{Average total token count across all benchmarks on Llama-3.1-8B-Instruct ($B=5$). Total tokens include output tokens and retrieved tokens.}
\label{fig:token_full}
\end{figure*}

\cref{fig:token_full} reports the average total token count across all seven benchmarks on Llama-3.1-8B-Instruct with $B=5$, where total tokens include both output tokens and retrieved tokens. Our method achieves the lowest token consumption on all datasets, confirming that budget-aware training reduces redundant search calls and the associated retrieval overhead.

\subsection{Budget Sampling Distribution}
\label{app:sampling}

\cref{tab:sampling_freq} shows how the budget sampling distribution evolves during Phase~II. At the onset of Phase~II (step 100), the distribution is uniform across all levels. As training progresses, the adaptive mechanism rapidly shifts probability mass toward lower budgets, which are inherently more challenging and exhibit lower accuracy. The distribution stabilizes after approximately 50 steps, with $B=1$ consistently receiving the highest sampling probability and $B=5$ the lowest. This confirms that the adaptive sampling mechanism successfully identifies and prioritizes the most difficult budget levels throughout training.

\begin{table}[t]
\centering
\small
\setlength{\tabcolsep}{7pt}
\caption{Average sampling probability for each budget level across training steps in Phase~II (Qwen3-4B).}

\label{tab:sampling_freq}
\begin{tabular}{lccccc}
\toprule
\rowcolor{gray!15}
\textbf{Step} & $B=1$ & $B=2$ & $B=3$ & $B=4$ & $B=5$ \\
\midrule
100 & 20\% & 20\% & 20\% & 20\% & 20\% \\
150 & 30\% & 24\% & 19\% & 15\% & 12\% \\
200 & 29\% & 22\% & 20\% & 17\% & 12\% \\
250 & 28\% & 23\% & 19\% & 18\% & 12\% \\
300 & 29\% & 22\% & 19\% & 17\% & 13\% \\
\bottomrule
\end{tabular}
\end{table}

\section{Cost Analysis}
\label{app:cost}

\begin{table*}[t]
\centering
\caption{Cost of one external search query (\$0.005) expressed as equivalent output tokens, based on OpenRouter pricing.}
\label{tab:cost_comparison}
\begin{tabular}{lccc}
\toprule
\rowcolor{gray!15}
\textbf{Model} & \textbf{Output Price} & \textbf{Cost per 1M Tokens} & \textbf{Equivalent Tokens} \\
\midrule
Qwen2.5-7B-Instruct   & \$0.10 / M tokens & \$0.10 & 50,000 \\
Qwen3-4B     & \$0.25 / M tokens & \$0.25 & 20,000 \\
LLaMA3.1-8B-Instruct  & \$0.05 / M tokens & \$0.05 & 100,000 \\
\bottomrule
\end{tabular}
\end{table*}

\begin{table}[t]
\centering
\caption{Token composition analysis on Llama-3.1-8B-Instruct ($B=5$). We report average total tokens, retrieval tokens, and the proportion of retrieval tokens in total tokens.}
\label{tab:token-ratio}
\resizebox{\linewidth}{!}{
\begin{tabular}{llccc}
\toprule
\rowcolor{gray!15}
\textbf{Dataset} & \textbf{Methods} & \textbf{Total} & \textbf{Retrieval} & \textbf{Ratio} \\
\midrule
\multirow{3}{*}{PopQA} & Search-R1  & 962 & 601 & 62.5\% \\
                       & StepSearch & 1263 & 728 & 57.6\% \\
                       & AnySearch & 895 & 594 & 66.4\% \\
\midrule
\multirow{3}{*}{2wiki} & Search-R1  & 1420 & 732 & 51.5\% \\
                           & StepSearch & 1527 & 1056 & 69.2\% \\
                           & AnySearch & 1215 & 784 & 64.5\% \\
\midrule
\multirow{3}{*}{Bamboogle} & Search-R1  & 2081 & 1410 & 67.8\% \\
                       & StepSearch & 1992 & 1161 & 58.3\% \\
                       & AnySearch & 1772 & 1150 & 64.9\% \\
\bottomrule
\end{tabular}
}
\end{table}

\subsection{Training Dynamics}
\label{app:training_dynamics}

\begin{figure*}[t]
\centering
\begin{subfigure}{0.3\linewidth}
\centering
\includegraphics[width=\linewidth]{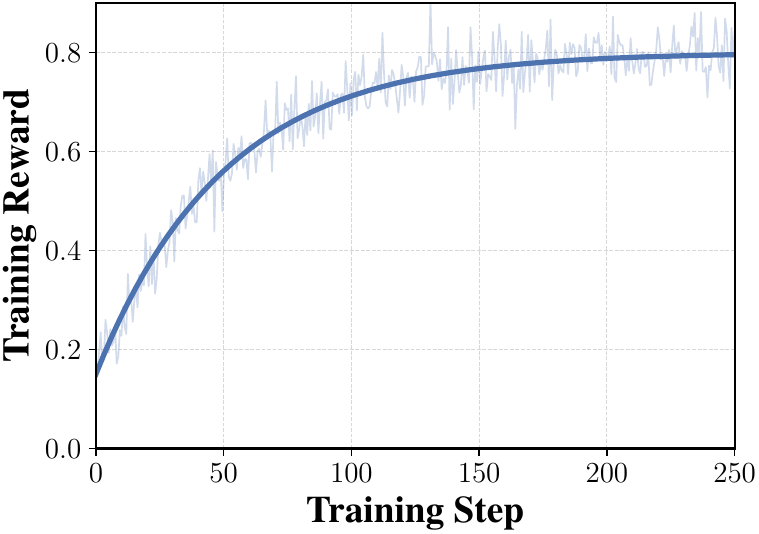}
\caption{Training Reward (Qwen2.5-7B)}
\end{subfigure}
\hfill
\begin{subfigure}{0.3\linewidth}
\centering
\includegraphics[width=\linewidth]{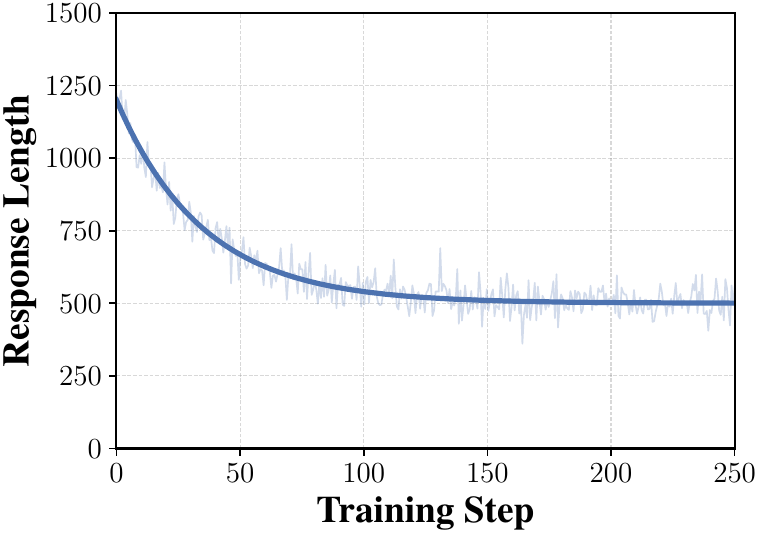}
\caption{Response Length (Qwen2.5-7B)}
\end{subfigure}
\hfill
\begin{subfigure}{0.3\linewidth}
\centering
\includegraphics[width=\linewidth]{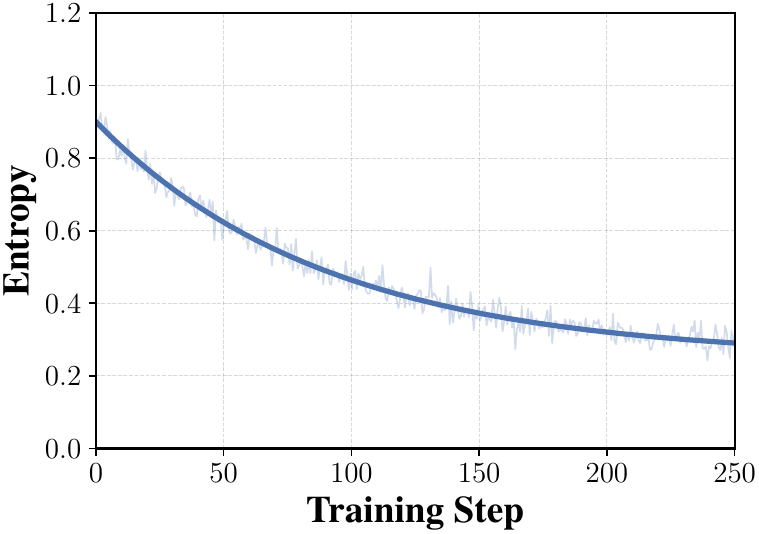}
\caption{Entropy (Qwen2.5-7B)}
\end{subfigure}

\vspace{0.3cm}

\begin{subfigure}{0.3\linewidth}
\centering
\includegraphics[width=\linewidth]{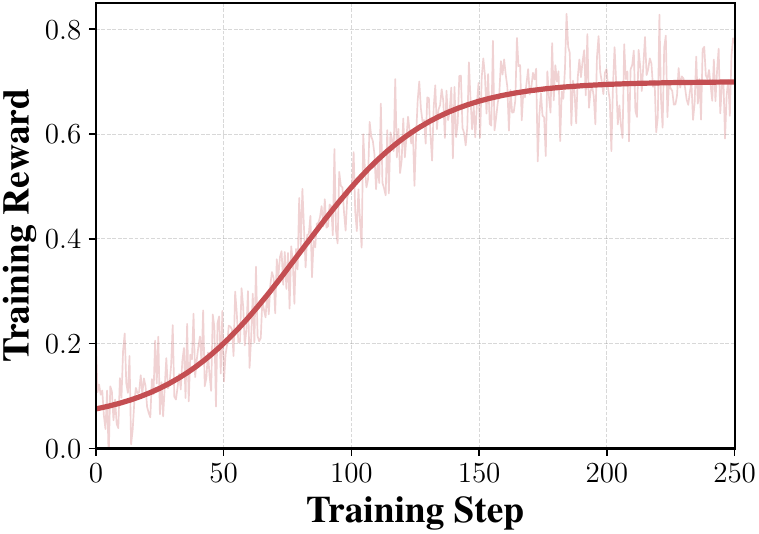}
\caption{Training Reward (Llama-3.1-8B)}
\end{subfigure}
\hfill
\begin{subfigure}{0.3\linewidth}
\centering
\includegraphics[width=\linewidth]{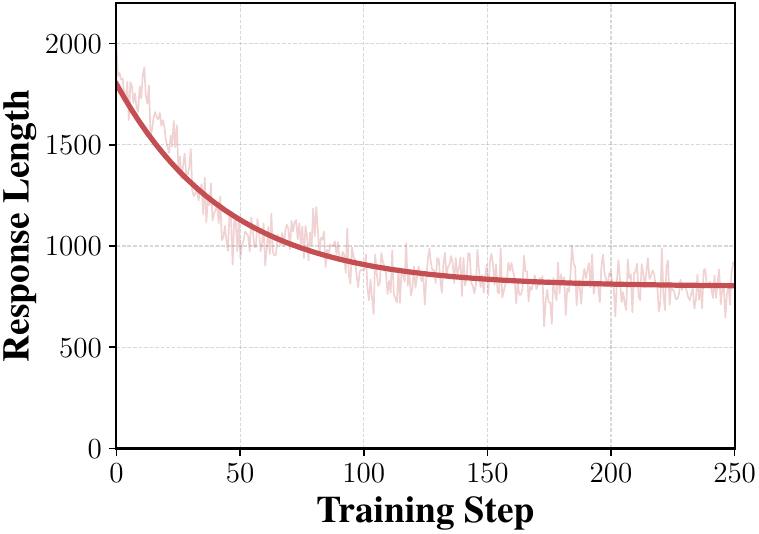}
\caption{Response Length(Llama-3.1-8B)}
\end{subfigure}
\hfill
\begin{subfigure}{0.3\linewidth}
\centering
\includegraphics[width=\linewidth]{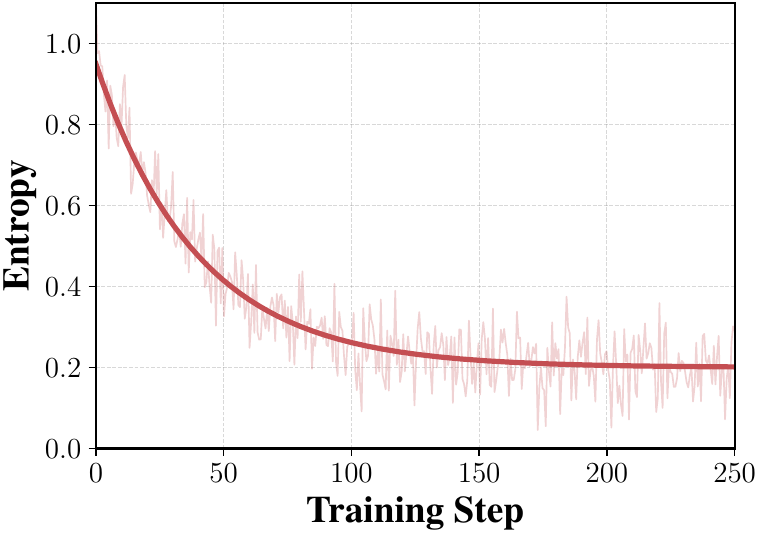}
\caption{Entropy (Llama-3.1-8B)}
\end{subfigure}
\caption{Training dynamics on Qwen2.5-7B-Instruct (top) and Llama-3.1-8B-Instruct (bottom), showing training reward, response length, and policy entropy across training steps.}
\label{fig:dynamic_other}
\end{figure*}

\cref{fig:dynamic_other} shows the training dynamics of Qwen2.5-7B-Instruct and Llama-3.1-8B-Instruct. Both models exhibit stable training curves consistent with the Qwen3-4B results reported in the main text: training reward increases steadily, response length remains within a reasonable range, and policy entropy decreases gradually without sudden collapse. These results confirm that our training framework generalizes stably across different model families and sizes.

\subsection{Datasets}
\label{app:datasets}

\paragraph{General QA Benchmarks.}
\begin{itemize}
    \item \textbf{Natural Questions (NQ):} Real anonymized Google search queries with Wikipedia-annotated answers. Tests open-domain QA grounded in a single reference page.
    \item \textbf{TriviaQA:} Trivia-style questions with independently collected evidence. Known for compositional questions and lexical mismatch requiring multi-sentence reasoning.
    \item \textbf{PopQA:} Long-tail factual QA derived from Wikidata triples. Probes reliance on parametric knowledge for low-popularity entities.
\end{itemize}

\paragraph{Multi-hop QA Benchmarks.}
\begin{itemize}
    \item \textbf{HotpotQA:} Distractor setting with gold and distractor paragraphs; evaluates multi-hop reasoning under noisy context.
    \item \textbf{2WikiMultiHopQA:} 2-hop compositional reasoning over Wikidata and Wikipedia with annotated evidence paths.
    \item \textbf{MuSiQue:} 2--4 hop questions with filtering for strict multi-hop requirement; tests evidence aggregation across multiple steps.
    \item \textbf{Bamboogle:} Manually written 2-hop questions designed to be difficult for keyword lookup; probes multi-step decomposition.
\end{itemize}

\subsection{Baselines Implementation}
\label{app:baselines}

\paragraph{Prompt-based Budget-Aware Baselines.}
\begin{itemize}
    \item \textbf{BATS~\cite{liu2025budget}:} Prompt-based budget-aware method that injects budget state via an external tracker at inference without RL training.
\end{itemize}

\paragraph{RAG-based Baselines.}
\begin{itemize}
    \item \textbf{Search-o1~\cite{li2025search}:} Inference-time agentic RAG interleaving reasoning with on-demand search queries without additional training.
\end{itemize}

\paragraph{RL-based Baselines.}
\begin{itemize}
    \item \textbf{Search-R1~\cite{jin2025search}:} RL-trained agent with outcome-based reward and retrieved-token masking.
    \item \textbf{ZeroSearch~\cite{sun2025zerosearch}:} RL with simulated retrieval; trains without live search engine calls.
    \item \textbf{StepSearch~\cite{wang2025stepsearch}:} Step-wise PPO~\cite{schulman2017proximal} with token-level information gain and redundancy penalties.
\end{itemize}

According to the pricing policy of the \texttt{Google Custom Search JSON API}\footnote{\url{https://developers.google.com/custom-search/v1/overview/}}, beyond the initial free tier, the service charges \$5.00 per 1,000 queries, resulting in a unit cost of \$0.005 per query.

In contrast, internal inference is significantly cheaper. As shown in Table~\ref{tab:cost_comparison}, the output token costs of the three backbone models (Qwen2.5-7B-Instruct, Qwen3-4B, and LLaMA3.1-8B-Instruct) used in our experiments are orders of magnitude lower than the per-query cost of external search, based on current market rates from \texttt{OpenRouter}\footnote{\url{https://openrouter.ai/}}. Even for the relatively more expensive Qwen3-4B, one external search query costs the equivalent of 20,000 output tokens. This stark disparity confirms that external search is the dominant cost factor in our agentic search systems, motivating our choice to define the budget constraint over external search calls.

Furthermore, \cref{tab:token-ratio} shows that retrieval tokens account for 51--68\% of total token consumption across all datasets in our experiments, confirming that the documents returned by search calls are the primary contributor to inference cost. This empirically validates that reducing unnecessary search calls is the most effective lever for lowering overall token overhead.

\section{Prompt Design}
\label{app:prompt}

The following prompt serves as the training scaffold used in Phase~I. It enforces a structured cycle of budget state observation, reasoning, and action, guiding the agent toward budget-aware decision patterns. This scaffold is removed in Phase~II and at inference.

\begin{promptbox}
You are an expert assistant designed to solve complex tasks through a rigorous cycle of reasoning and search. Your goal is to answer the user's question accurately while efficiently using your search budget. You have a total of {B} searches available. You must execute a structured loop of Budget, Reasoning, Action, Observation until you have sufficient information to answer.

1. Budget Phase (`<budget>`)
Before each reasoning step, you will see the current search budget:
<budget>remaining=R; used=U; total=T</budget>
The remaining is the number of searches you can still perform, used is the number of searches you have already performed, and total is the total budget allocated. Note that remaining + used = total.
You must consider the current budget state before deciding any action.

2. Reasoning Phase (`<think>`)
Before taking any action or providing an answer, you must output a reasoning block. Inside `<think>...</think>`, you are required to explicitly analyze:
Information Sufficiency: Is the current information adequate for producing a reliable answer? What specific information is still missing?
Budget Strategy: Analyze the current budget state (`<budget>`). Determine whether the next search is necessary and worthwhile given remaining resources. Prioritize using your internal knowledge when possible, and reserve search budget for information that is clearly beyond your parametric knowledge.

3. Search Phase (`<search>`)
If additional information is needed, invoke the search engine:
<search>query</search>
Cost: 1 unit of budget per search.
Usage: Use this when you need specific factual information that you cannot reliably produce from internal knowledge alone.

4. Observation Phase (`<information>`)
The system will return search results wrapped in `<information>...</information>` tags.

5. Answering Phase (`<answer>`)
Once you have gathered sufficient information and no further searching is required, provide the final concise answer wrapped in `<answer>...</answer>` tags. For example: <answer> Beijing </answer>

Now, answer the following question:
{Question}
\end{promptbox}

In Phase~II and at inference, the budget state injection and structured reasoning prompts are removed. The agent receives the following prompt, which specifies the total budget in natural language and retains only the basic interaction format:

\begin{promptbox}[title=\textbf{Phase~II / Inference Prompt}]
You are an expert assistant designed to solve complex tasks through reasoning and search. Your goal is to answer the user's question accurately. You have a total of {B} searches available. You must execute a loop of Reasoning, Action, Observation until you have sufficient information to answer.

1. Reasoning Phase (`<think>`)
Before taking any action or providing an answer, you must output a reasoning block inside `<think>...</think>` tags.

2. Search Phase (`<search>`)
If additional information is needed, invoke the search engine:
<search>query</search>
Cost: 1 unit of budget per search.

3. Observation Phase (`<information>`)
The system will return search results wrapped in `<information>...</information>` tags.

4. Answering Phase (`<answer>`)
Once you have gathered sufficient information and no further searching is required, provide the final concise answer wrapped in `<answer>...</answer>` tags. For example: <answer> Beijing </answer>

Now, answer the following question:
{Question}
\end{promptbox}

\section{Case Studies}
\label{app:cases}

We present five case studies on the same multi-hop questions. Cases 1--4 contrast Search-R1 and AnySearch under different budget constraints, illustrating that AnySearch internalizes budget-proportional allocation while Search-R1 lacks adaptive behavior regardless of budget availability. Case 5 further compares AnySearch with and without the training scaffold, demonstrating that the internalized policy preserves allocation quality while producing more concise reasoning.

\textbf{Case 1: Search-R1 @ Budget = 6 --- Premature Convergence under Abundant Budget.}
Despite having ample budget, Search-R1 terminates after only two searches and produces an incorrect answer (1631 instead of 1625). The agent retrieves the Dutch founding date of Philipsburg but fails to verify the earlier French arrival in the region. This reflects a key limitation observed in our main results (\cref{tab:exp_main}): without budget-aware training, the agent does not learn to leverage additional budget for verification, leading to premature convergence even when resources permit deeper exploration.

\begin{caseboxin}
Question: When did the people who captured Malakoff come to the region where Philipsburg is located?
Ground Truth: 1625

<think>I first need to identify who captured Malakoff.</think>
<search>who captured Malakoff</search>
<information>[...The Malakoff Redoubt was captured by French forces under General Pélissier during the Crimean War in 1855...]</information>

<think>The people were French. Now I need information about Philipsburg and its regional history.</think>
<search>Philipsburg history</search>
<information>[...Philipsburg is the capital of Sint Maarten, founded by the Dutch in 1631...]</information>

<think>The retrieved timeline gives 1631 for Philipsburg, which appears to answer the question.</think>
<answer>1631</answer>
\end{caseboxin}

\textbf{Case 2: AnySearch @ Budget = 6 --- Budget-Proportional Verification Depth.}
Given the same question and budget, AnySearch uses four searches to progressively decompose the multi-hop problem: identifying the French as the captors, locating Philipsburg, then verifying the earlier French settlement date. Crucially, the agent recognizes that the initial Dutch date (1631) is insufficient and invests additional budget to confirm the correct French arrival (1625). This demonstrates the budget-proportional behavior validated by \cref{fig:anybudget}: under abundant budgets, AnySearch deepens verification rather than stopping prematurely, yielding monotonically increasing accuracy with budget.

\begin{caseboxex}
Question: When did the people who captured Malakoff come to the region where Philipsburg is located?
Ground Truth: 1625

<think>I need to identify who captured Malakoff, then determine when those people arrived in the region where Philipsburg is located.</think>
<search>who captured Malakoff</search>
<information>[...The Malakoff Redoubt was captured by French forces under General Pélissier during the Crimean War in 1855...]</information>

<think>The people were French. I now need the relevant regional history connected to Philipsburg.</think>
<search>Philipsburg location history</search>
<information>[...Philipsburg is the capital of Sint Maarten, founded by the Dutch in 1631...]</information>

<think>The retrieved date refers to the Dutch founding of Philipsburg, but the question asks when the French came to the region. I should verify the earlier French presence in Saint Martin.</think>
<search>French settlement Saint Martin</search>
<information>[...French settlers established a presence in the St. Martin region by 1625, prior to Dutch colonization efforts...]</information>

<think>This gives an earlier French arrival date distinct from the later Dutch founding of Philipsburg. I will verify the timeline once more before answering.</think>
<search>Saint Martin French and Dutch colonization timeline</search>
<information>[...1625 marks the beginning of sustained French settlement in the Saint Martin area, while Philipsburg was founded later under Dutch rule...]</information>

<think>The evidence consistently indicates that the French arrived in the region in 1625.</think>

<answer>1625</answer>
\end{caseboxex}

\textbf{Case 3: Search-R1 @ Budget = 3 --- Poor Budget Allocation under Tight Constraints.}
Under a tight budget of 3, Search-R1 spends all three searches on this question but allocates them ineffectively. The agent retrieves Jonas Salk's biography and personal beliefs, obtaining conflicting information about his secular worldview versus Jewish heritage. Having exhausted its budget without resolving the ambiguity, it answers incorrectly. This exemplifies the efficiency gap measured by Tool Productivity (\cref{tab:effiency}): Search-R1 consumes more searches per correct answer because it lacks the ability to prioritize the most informative queries under constrained budgets.

\begin{caseboxinex}
Question: What was the religion of the inventor of the Polio vaccine?
Ground Truth: Jewish

<think>I need to identify the inventor of the Polio vaccine first.</think>
<search>history of the polio vaccine</search>
<information>[...Jonas Salk led the development of the first effective polio vaccine in 1955...]</information>

<think>The inventor was Jonas Salk. I should gather more background information before determining his religion.</think>
<search>Jonas Salk biography and beliefs</search>
<information>[...Jonas Salk described himself as a secular humanist and emphasized scientific rather than religious values...]</information>

<think>This gives his personal beliefs, but it is still unclear whether the question refers to religion, ethnicity, or upbringing. I need more clarification.</think>
<search>Jonas Salk family background</search>
<information>[...Jonas Salk was born to Jewish immigrant parents from Russia...]</information>

<think>I now have conflicting signals between his personal worldview and family background. Since no budget remains for further verification, I will answer based on his stated beliefs.</think>

<answer>Secular humanist</answer>
\end{caseboxinex}

\textbf{Case 4: AnySearch @ Budget = 3 --- Internalized Budget-Aware Allocation.}
Given the same question with only 3 searches, AnySearch allocates its limited budget with precision. The first search identifies Jonas Salk; the second directly targets his religious background, yielding the correct answer. The agent then reasons that the evidence is sufficient and terminates with one search remaining, rather than spending budget on redundant verification. This selective allocation---searching for the most discriminative information and stopping once confident---is the hallmark of the internalized budget-aware capability that our progressive scaffold removal produces, as confirmed by the superior TP scores in \cref{tab:effiency}.

\begin{caseboxinex2}
Question: What was the religion of the inventor of the Polio vaccine?
Ground Truth: Jewish

<think>I first need to identify the inventor of the Polio vaccine.</think>
<search>inventor of the polio vaccine</search>
<information>[...Jonas Salk developed the first effective polio vaccine in 1955...]</information>

<think>The inventor was Jonas Salk. I should directly verify his religious background.</think>
<search>Jonas Salk religious background</search>
<information>[...Jonas Salk was born to Jewish immigrant parents and was raised in a Jewish family...]</information>

<think>The retrieved information directly answers the question, so additional searches are unnecessary.</think>

<answer>Jewish</answer>
\end{caseboxinex2}

These four cases jointly illustrate the internalized budget-aware capability validated in our ablation studies (\cref{subfig:scaffold}): AnySearch adjusts its search depth proportionally to the available budget---investing in verification when resources are abundant (Case 2) and focusing on the most informative queries when resources are scarce (Case 4). In contrast, Search-R1 applies a fixed search strategy regardless of the budget, resulting in under-utilization under high budgets (Case 1) and inefficient allocation under low budgets (Case 3).

\textbf{Case 5: AnySearch with Scaffold @ Budget = 6 --- From Explicit Scaffold to Internalized Allocation.}
We additionally present the same question solved by AnySearch under the Phase~I scaffold (with explicit \texttt{<budget>} state injection and structured dual-factor reasoning). Comparing Case 5 with Case 2 reveals two important observations. First, both reach the correct answer with identical search trajectories (4 searches, same termination point), confirming the finding in \cref{subfig:scaffold} that Ours (+Scaffold) achieves comparable accuracy to Ours---the policy has fully internalized the budget-aware allocation that the scaffold teaches. Second, Case 2 (without scaffold) produces notably more concise reasoning: each \texttt{<think>} block focuses directly on the next action decision, whereas Case 5 (with scaffold) generates verbose Information Sufficiency and Budget Strategy analyses at every step. This difference is consistent with the lower total token consumption reported in \cref{fig:token}---after internalization, the agent compresses the explicit budget reasoning into implicit decision patterns, reducing thinking overhead while preserving the same allocation quality.

\begin{caseboxinex3}
% Case 5: AnySearch (Phase I Training Prompt with Scaffold) @ Budget = 6
Question: When did the people who captured Malakoff come to the region where Philipsburg is located?
Ground Truth: 1625

<budget>remaining=6; used=0; total=6</budget>
<think>Information Sufficiency: I do not yet know who captured Malakoff, so I cannot determine which people the question refers to. The missing information is the identity of the captors.
Budget Strategy: I have the full budget available. A search is necessary because the identity of the captors is specific factual information that should not be assumed from internal knowledge.</think>
<search>who captured Malakoff</search>
<information>[...The Malakoff Redoubt was captured by French forces under General Pélissier during the Crimean War in 1855...]</information>

<budget>remaining=5; used=1; total=6</budget>
<think>Information Sufficiency: I now know that the people were French. However, I still lack the relevant regional timeline connected to Philipsburg. I need information about where Philipsburg is located and the associated historical context.
Budget Strategy: With substantial budget remaining, an additional search is worthwhile. A targeted search about Philipsburg's location and history should efficiently reduce the remaining uncertainty.</think>
<search>Philipsburg location history</search>
<information>[...Philipsburg is the capital of Sint Maarten, founded by the Dutch in 1631...]</information>

<budget>remaining=4; used=2; total=6</budget>
<think>Information Sufficiency: I now have a date associated with Philipsburg, but the retrieved information describes Dutch founding activity. The question asks when the French came to the region where Philipsburg is located, so the missing information is the timeline of French presence in the Saint Martin region.
Budget Strategy: The current information is not yet reliable enough for answering. Since sufficient budget remains, a focused verification search about French settlement in Saint Martin is justified.</think>
<search>French settlement Saint Martin</search>
<information>[...French settlers established a presence in the St. Martin region by 1625, prior to Dutch colonization efforts...]</information>

<budget>remaining=3; used=3; total=6</budget>
<think>Information Sufficiency: I now have evidence for French presence in the region by 1625. This appears to answer the question, but I should confirm that this regional timeline is distinct from the later Dutch founding of Philipsburg.
Budget Strategy: I still retain ample remaining budget. One additional verification search is worthwhile to improve reliability before finalizing the answer.</think>
<search>French vs Dutch colonization Saint Martin timeline</search>
<information>[...1625 marks the beginning of sustained French settlement in the Lesser Antilles including the Saint Martin area. Philipsburg itself was founded later by the Dutch...]</information>

<budget>remaining=2; used=4; total=6</budget>
<think>Information Sufficiency: The retrieved evidence is now sufficient and internally consistent. The French arrived in the region in 1625, while Philipsburg was founded later by the Dutch. No critical information remains missing.
Budget Strategy: Further searching is unnecessary because the answer is supported by multiple consistent retrievals, and preserving remaining budget is preferable.</think>

<answer>1625</answer>
\end{caseboxinex3}

\subsection{Consolidating Evidence for Internalization}
\label{app:internal}

Beyond the ablation comparison in \cref{subfig:scaffold}, we consolidate three complementary signals from our experiments that jointly support the internalization claim.

\textbf{Scaffold redundancy at inference.} The Ours (+) variant in \cref{subfig:scaffold} reintroduces the full scaffold (budget state injection and structured reasoning prompts) at inference time. The resulting accuracy is virtually identical to our standard inference without the scaffold, indicating that the policy has already absorbed the budget-tracking capability the scaffold previously provided. If budget awareness were not internalized, re-enabling the scaffold should yield measurable gains.

\textbf{Active budget conservation.} The tool productivity results in \cref{tab:effiency} show that AnySearch uses substantially fewer searches per correct answer than baselines operating under the same budget ceiling. This implies that the policy actively conserves budget rather than defaulting to exhaustive search. A model without internal budget awareness would lack the basis for such selective allocation and would tend to use all available searches indiscriminately.

\textbf{Early stopping without external signals.} The case studies above demonstrate that AnySearch stops search early when confident. Case 4 uses only 2 of 3 available searches, and Case 2 stops at 4 of 6. This behavior requires implicit awareness of remaining budget without external state injection: the agent must simultaneously track that resources remain available and judge that further search is unnecessary. Search-R1, by contrast, either under-utilizes budget without deliberation (Case 1) or exhausts it without strategic allocation (Case 3).

Together, these three observations suggest that the trained policy maintains an internal representation of budget state and conditions its search decisions accordingly, rather than relying on external budget tracking at inference.

\begin{table*}[t]
\centering
\caption{Full comparison of efficiency across four multi-hop QA benchmarks under different backbone models and budget levels, measured by Accuracy (Acc) and Tool Productivity (TP).}
\label{tab:efficiency_full}
\resizebox{\textwidth}{!}{
\begin{tabular}{cl|cc|cc|cc|cc}
\toprule
\rowcolor{gray!15}
& & \multicolumn{2}{c|}{\textbf{HotpotQA}} & \multicolumn{2}{c|}{\textbf{2wiki}} & \multicolumn{2}{c|}{\textbf{MuSiQue}} & \multicolumn{2}{c}{\textbf{Bamboogle}} \\
\rowcolor{gray!15}
\textbf{Settings} & \textbf{Methods} & Acc$\uparrow$ & TP$\uparrow$ & Acc$\uparrow$ & TP$\uparrow$ & Acc$\uparrow$ & TP$\uparrow$ & Acc$\uparrow$ & TP$\uparrow$ \\
\midrule
\multicolumn{10}{l}{\textit{Qwen2.5-7B-Instruct}} \\
\midrule
\multirow{3}{*}{$B=4$} & \textbf{AnySearch} & \textbf{0.386} & \textbf{0.305} & \textbf{0.351} & \textbf{0.281} & \textbf{0.192} & \textbf{0.153} & \textbf{0.384} & \textbf{0.319} \\
& Search-R1 & 0.365 & 0.241 & 0.341 & 0.207 & 0.158 & 0.101 & 0.360 & 0.203 \\
& StepSearch & 0.374 & 0.223 & 0.330 & 0.189 & 0.172 & 0.118 & 0.312 & 0.179 \\
\midrule
\multirow{3}{*}{$B=5$} & \textbf{AnySearch} & \textbf{0.395} & \textbf{0.338} & \textbf{0.364} & \textbf{0.302} & \textbf{0.205} & \textbf{0.171} & \textbf{0.398} & \textbf{0.327} \\
& Search-R1 & 0.372 & 0.215 & 0.356 & 0.221 & 0.169 & 0.113 & 0.372 & 0.209 \\
& StepSearch & 0.384 & 0.258 & 0.342 & 0.196 & 0.185 & 0.107 & 0.328 & 0.218 \\
\midrule
\multirow{3}{*}{$B=6$} & \textbf{AnySearch} & \textbf{0.398} & \textbf{0.354} & \textbf{0.372} & \textbf{0.317} & \textbf{0.212} & \textbf{0.179} & \textbf{0.408} & \textbf{0.341} \\
& Search-R1 & 0.375 & 0.212 & 0.362 & 0.218 & 0.174 & 0.126 & 0.376 & 0.211 \\
& StepSearch & 0.388 & 0.276 & 0.348 & 0.203 & 0.190 & 0.108 & 0.336 & 0.194 \\
\midrule
\midrule
\multicolumn{10}{l}{\textit{Llama-3.1-8B-Instruct}} \\
\midrule
\multirow{3}{*}{$B=4$} & \textbf{AnySearch} & \textbf{0.394} & \textbf{0.347} & \textbf{0.378} & \textbf{0.309} & \textbf{0.190} & \textbf{0.158} & \textbf{0.400} & \textbf{0.338} \\
& Search-R1 & 0.356 & 0.249 & 0.335 & 0.218 & 0.176 & 0.119 & 0.352 & 0.231 \\
& StepSearch & 0.342 & 0.226 & 0.348 & 0.197 & 0.182 & 0.104 & 0.360 & 0.207 \\
\midrule
\multirow{3}{*}{$B=5$} & \textbf{AnySearch} & \textbf{0.402} & \textbf{0.378} & \textbf{0.392} & \textbf{0.334} & \textbf{0.203} & \textbf{0.168} & \textbf{0.412} & \textbf{0.351} \\
& Search-R1 & 0.364 & 0.236 & 0.349 & 0.213 & 0.188 & 0.122 & 0.365 & 0.227 \\
& StepSearch & 0.350 & 0.297 & 0.361 & 0.204 & 0.193 & 0.109 & 0.376 & 0.214 \\
\midrule
\multirow{3}{*}{$B=6$} & \textbf{AnySearch} & \textbf{0.408} & \textbf{0.386} & \textbf{0.398} & \textbf{0.349} & \textbf{0.208} & \textbf{0.177} & \textbf{0.416} & \textbf{0.362} \\
& Search-R1 & 0.368 & 0.231 & 0.354 & 0.208 & 0.192 & 0.131 & 0.368 & 0.219 \\
& StepSearch & 0.356 & 0.247 & 0.365 & 0.217 & 0.198 & 0.113 & 0.384 & 0.236 \\
\midrule
\midrule
\multicolumn{10}{l}{\textit{Qwen3-4B}} \\
\midrule
\multirow{3}{*}{$B=4$} & \textbf{AnySearch} & \textbf{0.378} & \textbf{0.309} & \textbf{0.362} & \textbf{0.291} & \textbf{0.152} & \textbf{0.124} & \textbf{0.336} & \textbf{0.274} \\
& Search-R1 & 0.351 & 0.235 & 0.340 & 0.216 & 0.124 & 0.093 & 0.328 & 0.201 \\
& StepSearch & 0.339 & 0.198 & 0.336 & 0.192 & 0.136 & 0.078 & 0.336 & 0.183 \\
\midrule
\multirow{3}{*}{$B=5$} & \textbf{AnySearch} & \textbf{0.384} & \textbf{0.321} & \textbf{0.375} & \textbf{0.308} & \textbf{0.164} & \textbf{0.137} & \textbf{0.352} & \textbf{0.283} \\
& Search-R1 & 0.356 & 0.231 & 0.353 & 0.204 & 0.132 & 0.088 & 0.344 & 0.206 \\
& StepSearch & 0.345 & 0.204 & 0.349 & 0.213 & 0.148 & 0.096 & 0.352 & 0.189 \\
\midrule
\multirow{3}{*}{$B=6$} & \textbf{AnySearch} & \textbf{0.388} & \textbf{0.332} & \textbf{0.380} & \textbf{0.313} & \textbf{0.170} & \textbf{0.143} & \textbf{0.360} & \textbf{0.291} \\
& Search-R1 & 0.360 & 0.218 & 0.358 & 0.213 & 0.138 & 0.097 & 0.352 & 0.204 \\
& StepSearch & 0.350 & 0.223 & 0.354 & 0.196 & 0.154 & 0.084 & 0.360 & 0.198 \\
\bottomrule
\end{tabular}
}
\end{table*}



\end{document}